\documentclass{article}

\usepackage[eandd, final]{neurips_2026}

\usepackage[utf8]{inputenc} %
\usepackage[T1]{fontenc}    %
\usepackage{hyperref}       %
\usepackage{url}            %
\usepackage{booktabs}       %
\usepackage{amsfonts}       %
\usepackage{nicefrac}       %
\usepackage{microtype}      %

\usepackage{amsmath}
\usepackage{amssymb}
\usepackage{longtable}
\usepackage{graphicx}
\usepackage{tabularx}
\usepackage{array}

\usepackage{multirow}
\usepackage{subcaption}

\usepackage[table]{xcolor}
\usepackage{makecell}
\usepackage{fvextra}
\usepackage[capitalize,nameinlink,noabbrev]{cleveref}
\crefname{section}{Section}{Sections}
\Crefname{section}{Section}{Sections}
\crefname{appendix}{Appendix}{Appendices}
\Crefname{appendix}{Appendix}{Appendices}

\title{Origami as a Real-Image Benchmark\\ for Procedural State-Transition Reasoning}

\author{%
  Ikumi Numaya$^{1}$ \quad
  Reina Akama$^{1,2,3}$ \quad
  Kise Yoshida$^{1}$ \quad
  Yuka Saito$^{1}$ \quad
  Sumika Tsuda$^{1}$ \\
  \textbf{Takahiro Shimizu$^{1}$} \quad
  \textbf{Miori Sagara$^{4}$} \quad
  \textbf{Jun Suzuki$^{1,3,5}$} \\
  $^{1}$Tohoku University \quad
  $^{2}$NINJAL \quad
  $^{3}$RIKEN \quad
  $^{4}$Baobab, Inc. \quad
  $^{5}$NII LLMC \\
  \texttt{is-failab-research@grp.tohoku.ac.jp} \\
}

\begin{document}

\maketitle

\begin{abstract}
General-purpose embodied AI systems are increasingly expected to interpret multimodal instructions and perform fine-grained physical tasks.
However, many manipulation benchmarks provide limited support for jointly evaluating procedural correctness, local geometric sensitivity, and human-guided correction.
We propose origami as a benchmark based on real images for procedural state-transition reasoning.
Origami requires precise operations such as tucking, inserting, reversing, and folding overlapping layers, making it a focused domain for studying physical state transitions.
We introduce \textsc{OrigamiTrace}\footnote{\url{https://tohoku-nlp.github.io/OrigamiTrace/}}, a dataset of step-by-step folding images, textual entries, and difficulty annotations, and define two tasks, i.e., autonomous state transitions from multimodal instructions and human-guided collaborative transitions.
Unlike crease-pattern- or diagram-based origami benchmarks, the proposed benchmark is grounded in photographs of actual intermediate folding states.
In an initial simulator-based evaluation, we use three multimodal LLM controllers to evaluate procedural understanding, action-interface grounding, and retry-based correction.
\end{abstract}

\section{Introduction}
As AI systems move beyond text and image processing, they are expected to operate as embodied agents that perceive, reason, and act in the physical world.
For example, in robotics, this shift appears in general-purpose robot policies, vision-language-action (VLA) models, and large-scale robot datasets that link perception, language, and action across a range of tasks and environments \cite{DBLP:conf/rss/BrohanBCCDFGHHH23,DBLP:conf/corl/ZitkovichYXXXXW23,DBLP:journals/corr/abs-2406-09246,DBLP:journals/corr/abs-2405-12213,DBLP:journals/corr/abs-2410-24164,DBLP:conf/icra/ONeillRMGPLPGMJ24,Khazatsky-RSS-24}.
However, evaluation is challenging for tasks requiring fine local geometry, long-horizon procedures, and human-guided correction.

\begin{figure}
  \centering
  \includegraphics[width=1.0\linewidth]{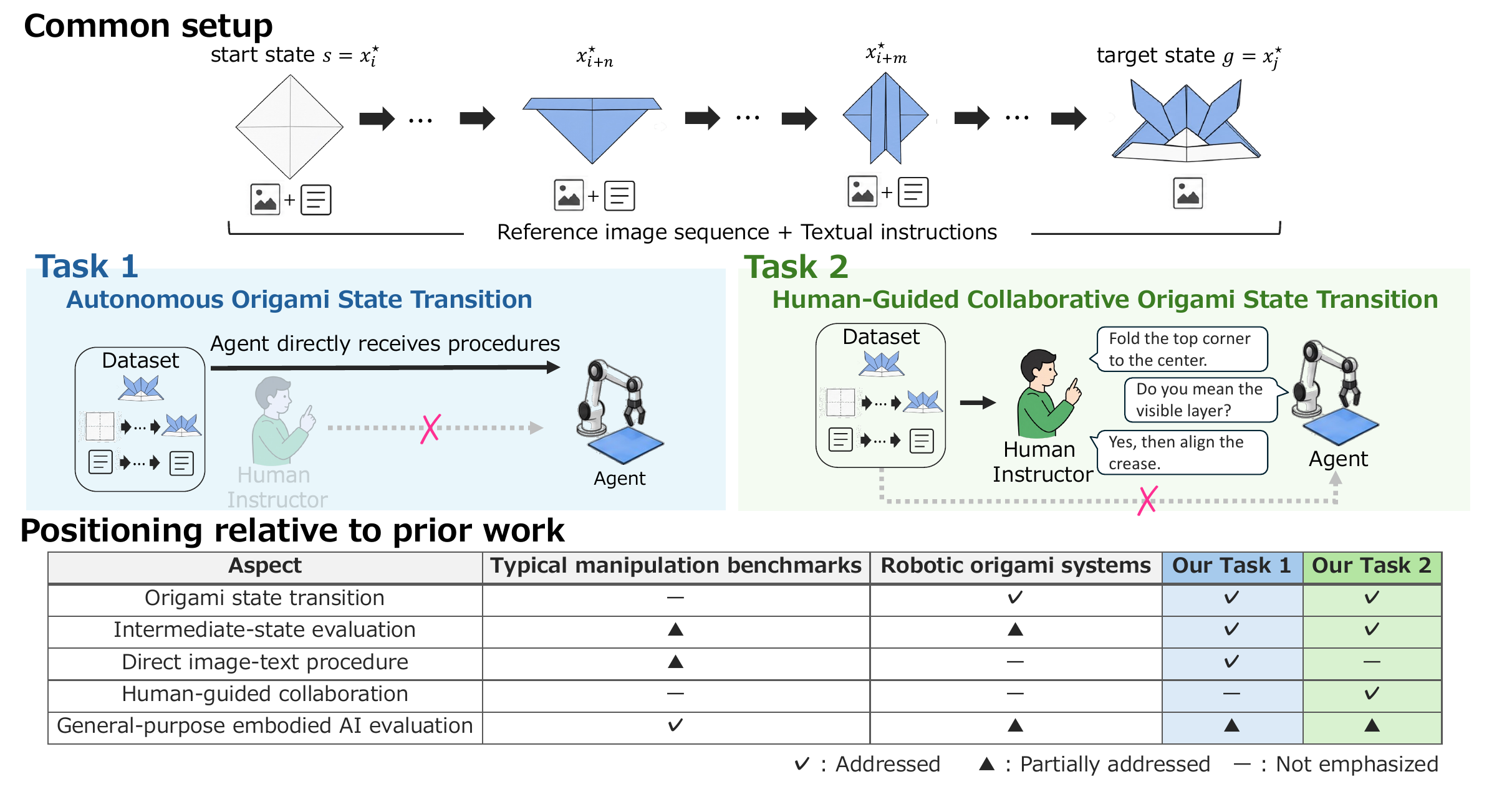}
  \caption{Overview of the proposed benchmark tasks and how they relate to typical manipulation and origami benchmarks.
A detailed comparison between Tasks 1 and 2 is provided in~\cref{tab:task_comparison_detail} in~\cref{appendix:task-comparison-detail}.
  }
  \label{fig:task}
\end{figure}

Existing manipulation benchmarks address multitask manipulation, language-conditioned control, long-horizon planning, and knowledge transfer \cite{DBLP:journals/ral/JamesMAD20,pmlr-v100-yu20a,DBLP:journals/corr/abs-2009-12293,DBLP:journals/ral/MeesHRB22,DBLP:conf/nips/MuLXYLLTHJ021,gu2023maniskill2,DBLP:conf/nips/LiuZGFLZS23,zhang2024vlabench}; however, many benchmarks emphasize task-level success, object configurations, or broad coverage across manipulation skills.
In this study, we focus on procedural state transitions, in which an agent must observe the current state, infer the intended next state, execute a local action, and adjust its behavior after failure.

We propose origami as a real-image-based benchmark for procedural state-transition reasoning.
Origami provides an inexpensive, easy-to-standardize domain that requires precise sequential operations involving crease placement, orientation, layer ordering, and hidden or overlapping reference points.
Differing from existing origami-inspired benchmarks that primarily evaluate spatial reasoning based on crease patterns, diagrams, or puzzle-like representations, the proposed benchmark treats origami as a procedural state-transition task grounded in real images of intermediate folded states.\footnote{We discuss this distinction in~\cref{sec:related-work} and~\cref{appendix:origami-positioning}.}

To support the proposed benchmark, we construct \textsc{OrigamiTrace}, a dataset of real images of step-by-step origami procedures paired with textual descriptions and difficulty annotations.
This dataset provides reference images of intermediate folded-paper states, allowing us to evaluate whether agents can ground visual and textual information in the current state and generate actions that advance the folding procedure.
We define two task settings, i.e., autonomous state transitions based on multimodal instructions and human-guided collaborative state transitions.

The proposed benchmark is intended to support future evaluations of simulators, physical robots, and VLA-based embodied systems; however, we do not claim to solve physical robotic origami manipulation.
For an initial simulator-based evaluation, we implement the tasks using multimodal LLMs that generate structured simulator commands.
The experiment evaluates procedure understanding, visual grounding, action-interface reasoning, and retry-based correction, rather than low-level motor control or full physical recovery from errors.
Extending the evaluation to physical execution is expected to better align the proposed benchmark with real-world conditions.

The primary contributions of this study are summarized as follows.
First, we formulate origami as a benchmark domain for procedural state-transition reasoning.
Second, we construct \textsc{OrigamiTrace}, a real-image origami dataset with textual descriptions, difficulty annotations, and validation of final-state plausibility.
Third, we define autonomous and human-guided state-transition tasks and implement them in a simulator using multimodal LLM controllers.

\section{Task definitions}
\label{sec:Task-Definition}

This paper defines two origami state-transition tasks (\cref{fig:task}).
Task 1 evaluates whether an agent can follow a multimodal origami procedure given as reference images and textual instructions.
Task 2 evaluates whether an agent can achieve the state-transition objective through step-by-step guidance from a human instructor with access to the procedure.
These tasks distinguish direct procedure following from instructor-mediated execution.
This distinction is important because embodied agents may not always receive complete task specifications directly. Instead, they may need to act from partial instructions and refine their behavior through interactive or corrective human guidance.\footnote{\cref{appendix:origami-motivation} gives additional motivation for this task formulation.}

An \emph{agent} may be a physical robot, a VLA-based policy, a simulated embodied system, or an LLM-based controller operating through a simulator interface.
Thus, actions may correspond to physical robot motions, manipulation primitives, or structured simulator commands.
In this study, as an initial evaluation, we employ a multimodal LLM that operates a simulator through commands as the agent.

\subsection{Common setup}

Let \(x_t \in \mathcal{X}\) denote the underlying paper state at step \(t\), including geometry, crease structure, orientation, layer ordering, and contact conditions.
The agent receives an observation \(o_t \in \mathcal{O}\), such as RGB images, RGB-D images, point clouds, or rendered simulator observations.
Note that origami frequently contains overlapping and hidden layers; thus \(o_t\) may only partially reveal \(x_t\).
An action \(a_t \in \mathcal{A}\) changes the state according to
\(
x_{t+1}=F(x_t,a_t),
\)
where \(F\) is induced by the simulator in simulation or by physical interaction in the real world.
A benchmark instance is denoted \(q=(s,g,I)\), where \(s\), \(g\), and \(I=(I^{\mathrm{img}},I^{\mathrm{text}})\) represent the start state, the target state, and the reference procedure, respectively.
The image sequence \(I^{\mathrm{img}}=(v_0,\ldots,v_K)\) depicts reference states \(x_0^\star,\ldots,x_K^\star\), and the textual instructions \(I^{\mathrm{text}}=(u_0,\ldots,u_{K-1})\) describe the transitions between adjacent states.
For a sampled episode, the start and target states correspond to an ordered pair \(s=x_i^\star\) and \(g=x_j^\star\), where \(0 \leq i < j \leq K\).
Therefore, an instance may be a single fold, a multistep segment, or a full origami procedure.
In this process, \(h_t=(o_0,a_0,o_1,\ldots,a_{t-1},o_t)\) is the agent's interaction history.
For interactive settings, \(d_t\) denotes the dialogue history.
A noninteractive policy outputs \(a_t \sim \pi(\cdot \mid h_t,\kappa)\), where \(\kappa\) denotes the task information that is directly available to the agent.
Furthermore, an interactive policy is conditioned on the dialogue history \(d_t\) and may output either an action \(a_t\) or a clarification question \(r_t \in \mathcal{R}\).
For Task 1, the task information is \(\kappa=(g,I)\).
For Task 2, \(\kappa\) does not include direct access to \(g\), \(I^{\mathrm{img}}\), or \(I^{\mathrm{text}}\). This information is only available through the human instructor.
When a human instructor is present, they may communicate with the agent but may not physically manipulate the paper or directly operate the environment.

\subsection{Evaluation philosophy}
\label{subsec:evaluation-philosophy}

The agent's origami performance is evaluated using multiple separate metrics rather than a single linearly combined score.
The primary metric is the \emph{episode success rate}, indicating whether the agent completes all required transitions from the sampled start state to the target state within the permitted number of attempts.
This metric directly measures the completion of the state-transition objective.
Because origami procedures can be lengthy and errors may accumulate, episode success rate may be too sparse at low performance levels.
Therefore, we also consider several transition-level metrics, including the \emph{transition success rate}, \emph{first-attempt success rate}, and \emph{average attempts per transition}, that diagnose partial progress and local failure modes.
In origami, these diagnostics are important because final-state success alone does not show whether the agent recognized the current state, selected the correct layer, executed the intended local operation, or made an early error that propagated to subsequent steps.
Furthermore, we evaluate correction behavior separately.
A full recovery evaluation measures whether an agent can realign with the reference state sequence after perturbations or execution errors.
When an environment does not support reliable corrective manipulation from arbitrary folded states that deviate from the reference, retry-based correction can function as a practical proxy for effective recovery behavior.
Retry-based correction evaluates whether an agent can utilize failure feedback to revise its action choice rather than merely producing a single open-loop command.
Additional motivation for retry-based correction is discussed in~\cref{appendix:origami-motivation}.

\subsection{Task 1: Autonomous origami state transition from multimodal instructions}

In Task 1, the agent is given the target state \(g\) and the reference procedure \(I=(I^{\mathrm{img}},I^{\mathrm{text}})\).
The policy is denoted \(a_t \sim \pi_1(\cdot \mid h_t,g,I)\), and no instructor guidance is permitted.
The agent must match its current observation with the reference procedure, infer the next transition from the image-text instructions, and execute the state change.
This task tests whether the agent can follow multimodal procedures directly. The agent can access the local reference procedure; however, it must still ground the visual and textual information in the current paper state and produce an executable action.

Task 1 evaluates multimodal procedure understanding, visual-language grounding, state recognition, procedural execution, and fine-grained state-transition reasoning.
The primary metric is episode success rate.
We also report transition-level metrics and retry-correction success rate, which is defined as the fraction of failed first attempts that are corrected within the remaining retry budget after receiving fixed feedback.
This metric quantifies retry-based correction and approximates recovery-related behavior, but does not measure full recovery from a physically deviated paper state.

\subsection{Task 2: Human-guided collaborative origami state transition}

Task 2 uses the same benchmark instance \(q=(s,g,I)\), but the agent is not given \(g\), \(I^{\mathrm{img}}\), or \(I^{\mathrm{text}}\) directly.
Instead, a human instructor observes the reference procedure and guides the agent through dialogue.
Here, the interactive policy, denoted \(\pi_2(\cdot \mid h_t,\kappa,d_t)\), may output either an action \(a_t\) or a clarification question \(r_t \in \mathcal{R}\).
The dialogue history \(d_t\) contains instructor instructions, agent questions, instructor answers, and corrective feedback.

Task 2 evaluates instructor-mediated state-transition reasoning.
In this task, the agent must act without direct access to the full reference procedure. The agent uses only its current observation, the dialogue history, and the information conveyed by the instructor.
This task setting captures collaborative embodied scenarios in which a human user may guide an agent using partial instructions, local clarification, or corrective feedback rather than providing a complete formal task specification.

In this context, the instructor may read, paraphrase, or locally clarify the relevant reference instruction, answer clarification questions, and provide corrective feedback.
However, they cannot physically manipulate the paper, directly specify the action command to the agent, or disclose the full future reference sequence in advance.
The objective is to make the agent's paper state follow the reference state sequence rather than reproduce the reference text. In this task, the human instructor provides the guidance; thus the agent's performance can vary in response to different factors, including the instructor's phrasing and clarification behavior.

As in Task 1, the primary metric is episode success rate, with transition-level metrics used for diagnosis.
Retry-based correction in Task 2 is human-guided. The instructor provides corrective feedback after a failed attempt, and the agent attempts the corresponding transition again.
In addition, we record various interaction statistics, including instructor turns, clarification questions, feedback turns, and instruction length.

\subsection{Recovery evaluation mode}

The recovery mode is an optional extension of both Tasks 1 and 2 that evaluates whether the agent can realign with the reference state sequence following perturbations, execution errors, or ambiguous intermediate states.
In Tasks 1 and 2, recovery is autonomous and human-guided through corrective dialogue, respectively. 
The recovery mode is important because small local errors in origami can persist through later folds, potentially affecting geometry, visibility, and layer ordering.

To perform a full recovery evaluation, the environment must preserve deviated paper states and allow corrective actions from those states.
However, the simulator employed in the initial experiments did not yet reliably support this; thus, \cref{sec:Benchmark-Simulation} evaluates retry-based correction, where a human evaluator judges command correctness and the simulator keeps the current state image unchanged after an unsuccessful attempt. 
The agent then retries the same transition after receiving feedback.
This approximates recovery-related behavior, but it does not represent a complete evaluation of recovery from a physically deviated paper state.

\section{Related work}
\label{sec:related-work}

\paragraph{Embodied manipulation benchmarks.}
General-purpose robot policies, VLA models, large-scale robot datasets, and manipulation benchmarks have driven progress in language-conditioned control, long-horizon planning, and knowledge transfer across a wide range of tasks and embodiments
\cite{DBLP:conf/rss/BrohanBCCDFGHHH23,DBLP:conf/corl/ZitkovichYXXXXW23,DBLP:journals/corr/abs-2406-09246,DBLP:journals/corr/abs-2405-12213,DBLP:journals/corr/abs-2410-24164,DBLP:conf/icra/ONeillRMGPLPGMJ24,Khazatsky-RSS-24,DBLP:journals/ral/JamesMAD20,pmlr-v100-yu20a,DBLP:journals/corr/abs-2009-12293,DBLP:journals/ral/MeesHRB22,DBLP:conf/nips/MuLXYLLTHJ021,gu2023maniskill2,DBLP:conf/nips/LiuZGFLZS23,zhang2024vlabench,pmlr-v202-jiang23b}.
The proposed benchmark complements this line of research by focusing on a compact procedural domain in which local errors, hidden or overlapping layers, the order of actions, and correction behavior shape later state transitions.

\paragraph{Fine-grained, deformable, and paper manipulation.}
Origami overlaps with fine-grained, bimanual, and deformable-object manipulation, including studies of cloth, cables, bags, differentiable simulation, sim-to-real gaps, and language-guided folding
\cite{DBLP:journals/corr/abs-2304-13705,pmlr-v270-fu25b,DBLP:conf/rss/ChiFDXCBS23,pmlr-v164-mandlekar22a,pmlr-v155-lin21a,chen2023daxbench,DBLP:conf/icra/SeitaFTCSGZ21,DBLP:journals/ral/MuleroBACTK24,osti_10345700,DBLP:journals/corr/abs-2508-16749,DBLP:journals/arcras/LonghiniWGBMWAYEHBK25,bifold}.
Previous studies on robotic origami and paper folding have also explored specialized mechanisms, controllers, and physical folding strategies
\cite{DBLP:conf/icra/BalkcomM04,1360855568490919680,1360013199658013056,DBLP:conf/humanoids/ElbrechterHR12,DBLP:conf/iros/NamikiY15,DBLP:conf/robio/NakashimaIH16,kitamori2024origami,arasaki2024realtime}.
In contrast, the goal of the current study is to define origami as a benchmark domain for state-transition reasoning, procedural understanding, and correction behavior in simulated settings and future embodied settings.

\paragraph{Origami-inspired reasoning benchmarks.}
Recently, origami-inspired benchmarks have been developed to evaluate spatial reasoning using crease patterns, diagrams, mathematical constraints, puzzle-like representations, or programmatic folding actions
\cite{xu2025neurips-origamispace,spencer2025gamibenchevaluatingspatialreasoning,paperfolding2026benchmark,agarwal2026can}.
These benchmarks are valuable for geometric reasoning and 2D--3D spatial reasoning. However, the proposed benchmark addresses a different problem, i.e., reasoning from actual intermediate folded states to the next intended state.
Further discussion is provided in~\cref{appendix:origami-positioning}.

\paragraph{Positioning.}
The proposed benchmark lies between origami-based spatial reasoning benchmarks and embodied manipulation benchmarks.
The novelty of our benchmark lies in the use of origami and in framing origami as a procedural state-transition benchmark based on real images, with task formats and metrics intended for simulated and future physical agents.
Further motivation is provided in~\cref{appendix:motivation-procedural-task,appendix:motivation-procedural-domain}.

\begin{figure}[t]
  \centering
  \includegraphics[width=\linewidth]{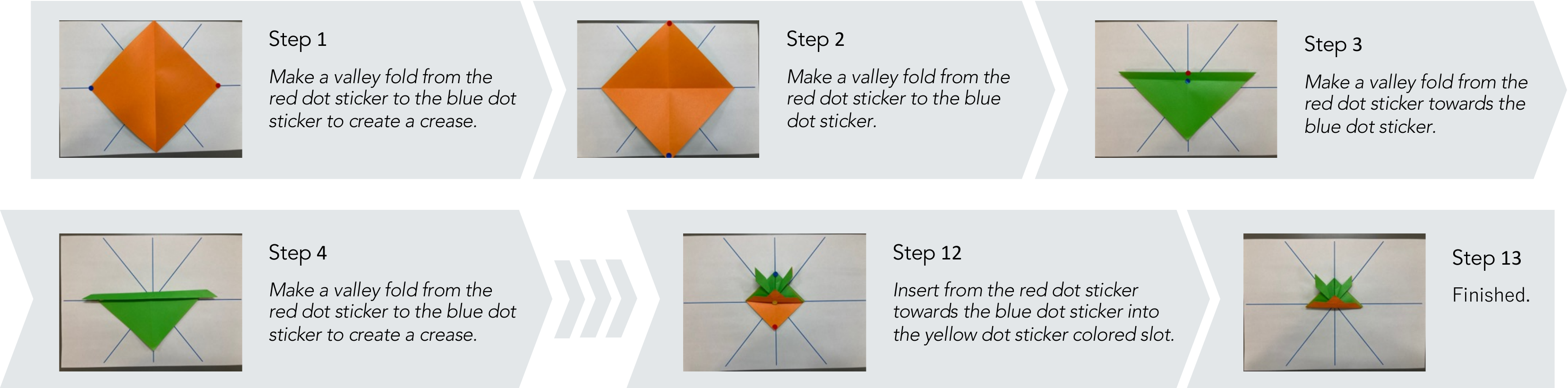}
  \caption{Example from the dataset: ``Tsunonaga-Kabuto'' (long-horned samurai helmet).}
  \label{fig:dataset}
\end{figure}

\section{Dataset construction}
\label{sec:dataset_construction}

\subsection{Dataset overview}

Our benchmark dataset, \textsc{OrigamiTrace}, contains 348 origami designs, comprising a total of 7,221 captured images corresponding to reference paper states.
Each captured image is paired with a textual entry, yielding a total of 7,221 image--text pairs.
For each nonfinal step, the textual entry describes the operation required to transition to the next state. For the final step, the textual entry marks completion, e.g., \textit{Finished}.
A sample from the dataset is shown in \cref{fig:dataset}.
Each design is represented by a sequence of state images, beginning with the paper preparation (Step 0: \textit{Prepare origami paper}) and continuing through to completion.
Difficulty levels are assigned to all designs on a five-point scale, based on the complexity of the folding operations and the number of steps involved.

The dataset comprises real images and corresponding human-written captions captured in a controlled, fiducial-assisted setup rather than unconstrained natural images. 
The images and textual entries were created manually under a standardized worker-facing specification and reviewed through a multi-stage quality-control workflow. 
During the data creation process, checkers reviewed the images and textual entries against the worker-facing specification; details are given in~\cref{appendix:worker-facing-specification}.
After data creation, each transition was evaluated from three predefined perspectives, ensuring that it was foldable by a person following the written instruction; details are given in~\cref{subsec:dataset_construction_procedure}.

For benchmarking, each design is treated as one reference procedure \(I\) in~\cref{sec:Task-Definition}.
A benchmark episode is formed by selecting a start state and a later target state from the same procedure, together with the corresponding state images and textual entries.
In Task 1, the selected state images and textual information are provided directly to the agent.
In contrast, in Task 2, the same reference procedure is available to the human instructor, while the agent only receives the current simulator observation and instructor guidance.
Thus, a single dataset supports both direct multimodal procedure following and instructor-mediated state-transition reasoning.

\subsection{Construction procedure}
\label{subsec:dataset_construction_procedure}

\paragraph{Collecting reference origami designs.}
To compile a list of origami designs to be included in the dataset, we collected information from websites that provide folding instructions for early childhood education and recreational use.
Because the benchmark is intended to support future evaluation of embodied agents and robotic systems, we excluded designs that require tools or operations beyond conventional paper manipulation, e.g., cutting with scissors, using adhesives, or combining multiple independent pieces of paper.
Ultimately, the dataset contained 348 unique designs.

\paragraph{Capture of procedural steps.}
Given the reference origami designs, workers reproduced the corresponding folding processes and captured images of each step until completion.
These images were captured in a controlled visual setup rather than in unconstrained natural scenes.
We used a white background with blue radial guide lines extending in eight directions at 45° intervals from the center, thereby providing effective visual references for orientation, rotation, and placement.
The origami paper was green on one side and orange on the other to facilitate boundary detection and observation of surface reversals \cite{tpami2011arbelaez,iccv2015fu}.
In some steps, small dot stickers were attached to the paper or background to indicate key vertices or reference positions for folding.
These stickers serve as fiducial cues for instruction captioning and state grounding.
During the image capture process, the origami was positioned at the center of the radial pattern and photographed from a top-down perspective.

\paragraph{Instruction captioning.}
A textual entry was manually assigned to each captured image.
For the nonfinal steps, the textual entry describes the folding operation required to reach the next state, specifying which part to fold, in which direction, and how.
For the final steps, the textual entry marks completion (e.g., \textit{Finished}).
To ensure consistent terminology across the dataset, a set of fundamental folding operations (e.g., \textit{mountain fold, valley fold, and turn over}) and commonly used techniques (e.g., \textit{squash fold, insert}) was predefined and provided to the annotators.
In this process, the dot stickers were used to mark positions and directions, and they were referenced as required during the instruction creation process (e.g., \textit{Make a valley fold from the red \textbf{dot sticker} to the blue \textbf{dot sticker}}).
Note that these stickers were occasionally explicitly referenced in instructions. Thus, the current dataset should be considered a controlled, fiducial-assisted origami-procedure dataset rather than an unconstrained natural-image dataset.

\paragraph{Quality checking.}
All image--text pairs were manually assessed in terms of the quality of each transition. 
For each transition, the checkers compared the images of the two adjacent paper states with the corresponding textual entry, and then they assessed the transition from three perspectives, i.e., foldability, specification conformance, and text quality.
In terms of foldability, the checkers verified whether a person could actually fold the paper from the given state to the next state by following the textual entry.
The checkers also verified that the textual entry described the intended transition between the adjacent paper states (e.g., whether the folded layers, rotations, and turn-overs were consistent with the images).
In terms of specification conformance, the checkers verified that the entry followed the predefined terminology and referred to visible fiducial cues consistently when necessary.
In terms of text quality, the checkers examined the entries for fine-grained errors, e.g., typos and inconsistent wording.
For transitions with identified issues, the image or the textual entry was corrected following the same procedure used in the original creation process, and then a different checker independently reviewed the transition from the same three perspectives.
This correction-and-recheck cycle was repeated until no issues remained relative to the three perspectives, at which point the data for the transition were finalized.
Additional details are provided in~\cref{appendix:transition-quality-control}.

\paragraph{Final-state plausibility evaluation.}
The final-state images were evaluated through crowdsourcing to assess whether they were visually recognizable as their assigned labels.
The crowdworkers were shown the image of the final step together with a label, i.e., the name of the origami design, and they answered \textit{yes} or \textit{no} to indicate whether the label described the image correctly.
Two settings were used in this evaluation.
In the true-label setting, the correct label was paired with the image, and the expected answer was \textit{yes}.
In the false-label setting, an incorrect label was selected randomly and paired with the image, and the expected answer was \textit{no}.
Both settings were applied to all 348 designs, and each item was evaluated by six crowdworkers (three English speakers and three Japanese speakers).
The overall label-decision accuracy was 83.7\%, providing a coarse check of final-state plausibility.
Inter-annotator agreement was higher among the Japanese speakers ($\kappa=0.71$) than among the English speakers ($\kappa=0.54$).
Additional details are provided in~\cref{appendix:dataset-quality-evaluation}.

\begin{table}[t]
    \centering
    \caption{Distribution of difficulty levels of origami designs in the dataset.}
    \label{tab:difficulty}
    \small
    \renewcommand{\arraystretch}{1.2}
    \begin{tabular}{lp{7cm}r}
        \toprule
        Difficulty
        & Criteria
        & Frequency (\%) \\
        \midrule
        Level 1 -- Beginner
        & \footnotesize{Approximately 1--5 steps; simple designs that can be completed using basic folding techniques.}
        & 58 (16.7\%) \\
        \rowcolor{gray!10}
        Level 2 -- Elementary
        & \footnotesize{Approximately 5--10 steps; typically comprising relatively flat two-dimensional designs.}
        & 131 (37.6\%) \\
        Level 3 -- Intermediate
        & \footnotesize{Approximately 10--50 steps; includes three-dimensional forms and moderately complex patterns, requiring sustained attention.}
        & 144 (41.4\%) \\
        \rowcolor{gray!10}
        Level 4 -- Advanced
        & \footnotesize{Greater than 50 steps; characterized by numerous intricate folds and longer completion times, frequently involving variant techniques.}
        & 15 (\; 4.3\%) \\
        Level 5 -- Expert
        & \footnotesize{Greater than 400 steps; extremely advanced designs.}
        & 0 (\; 0.0\%) \\
        \midrule
        Total
        &
        & 348 (100.0\%) \\
        \bottomrule
    \end{tabular}
\end{table}

\begin{figure}[t]
    \centering
    \tabcolsep 1pt
    \small
    \begin{tabular}{cccc}
    \includegraphics[width=0.22\linewidth]{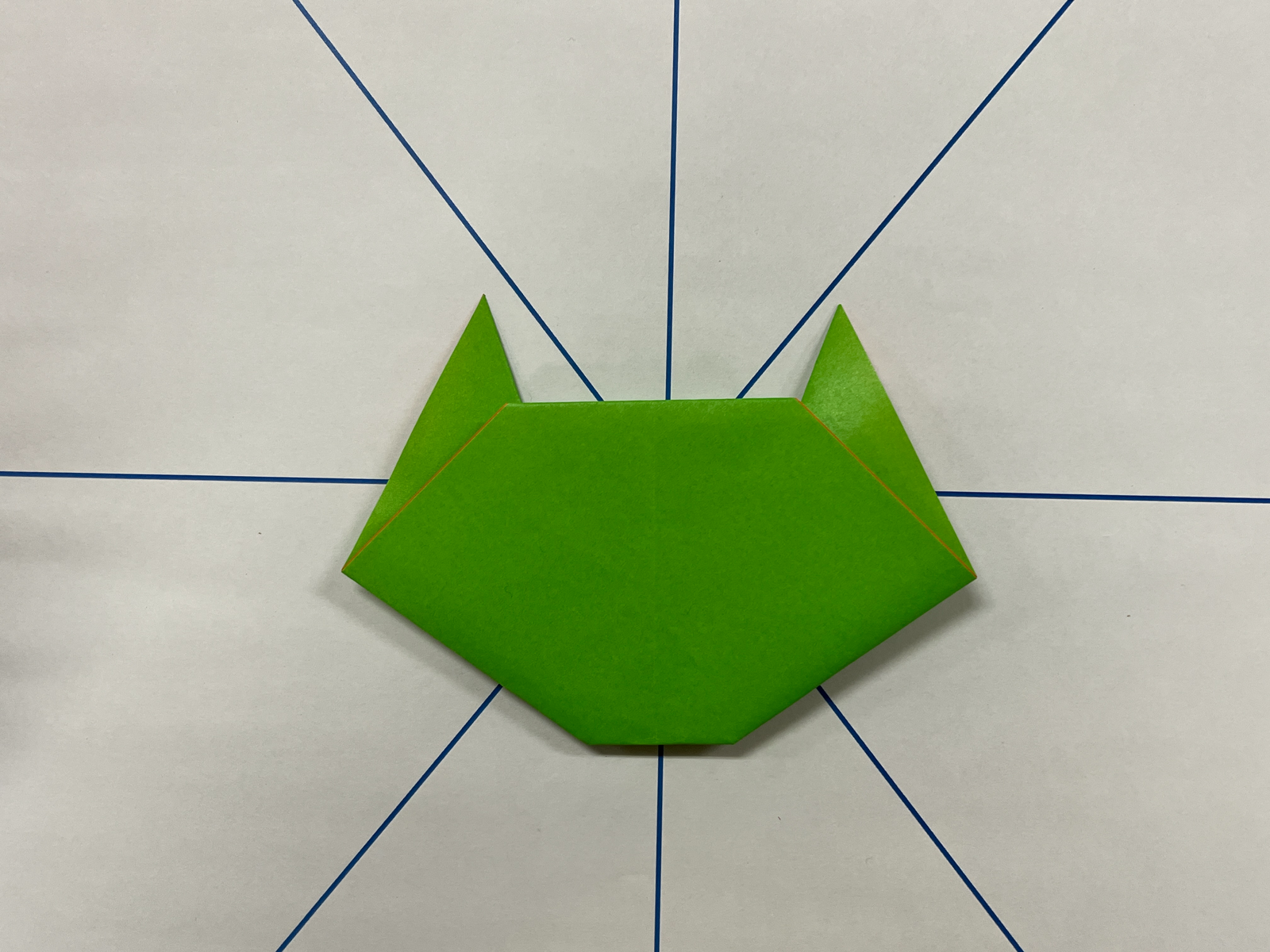}
&    \includegraphics[width=0.22\linewidth]{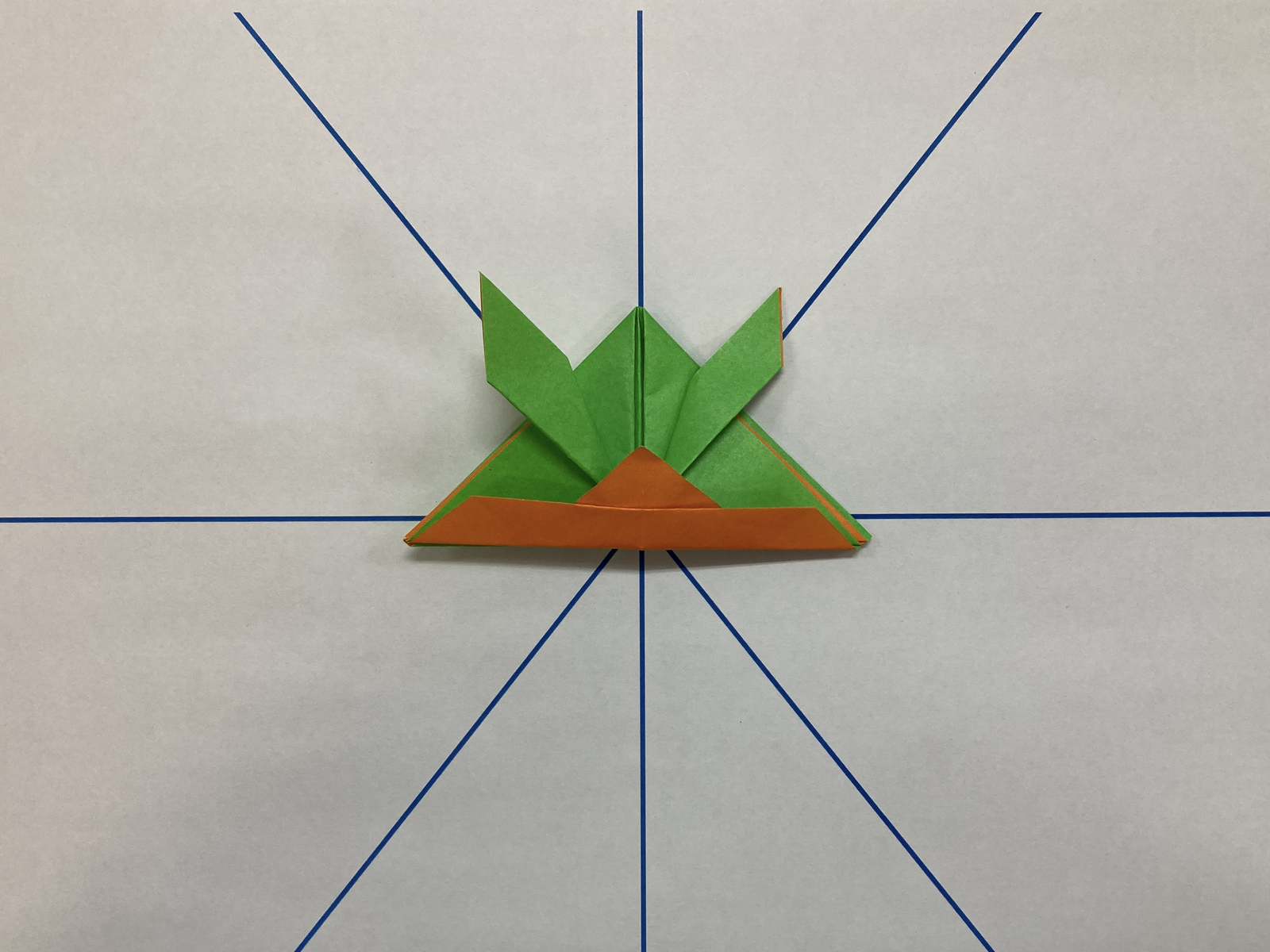}
&    \includegraphics[width=0.22\linewidth]{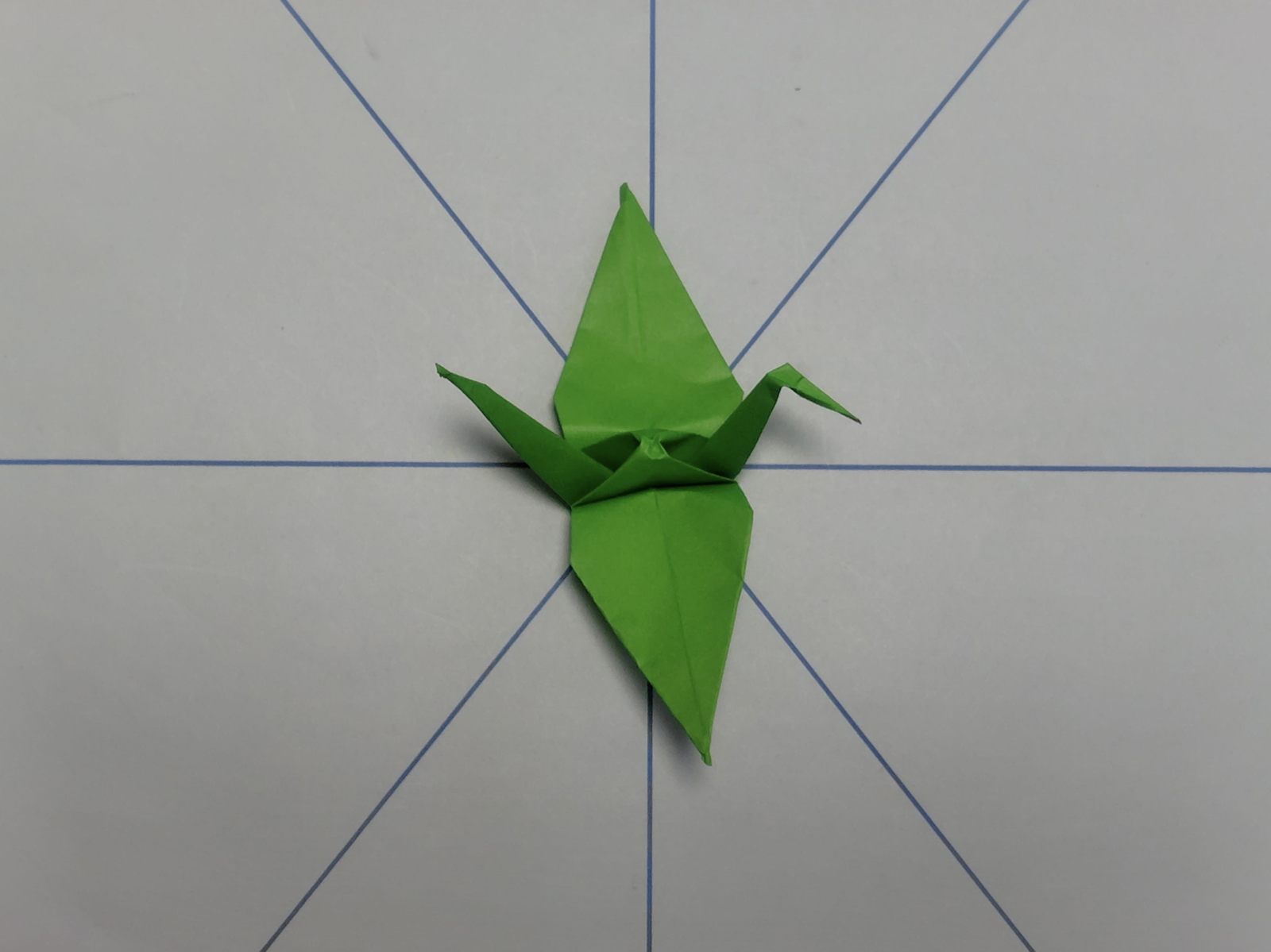}
&    \includegraphics[width=0.22\linewidth]{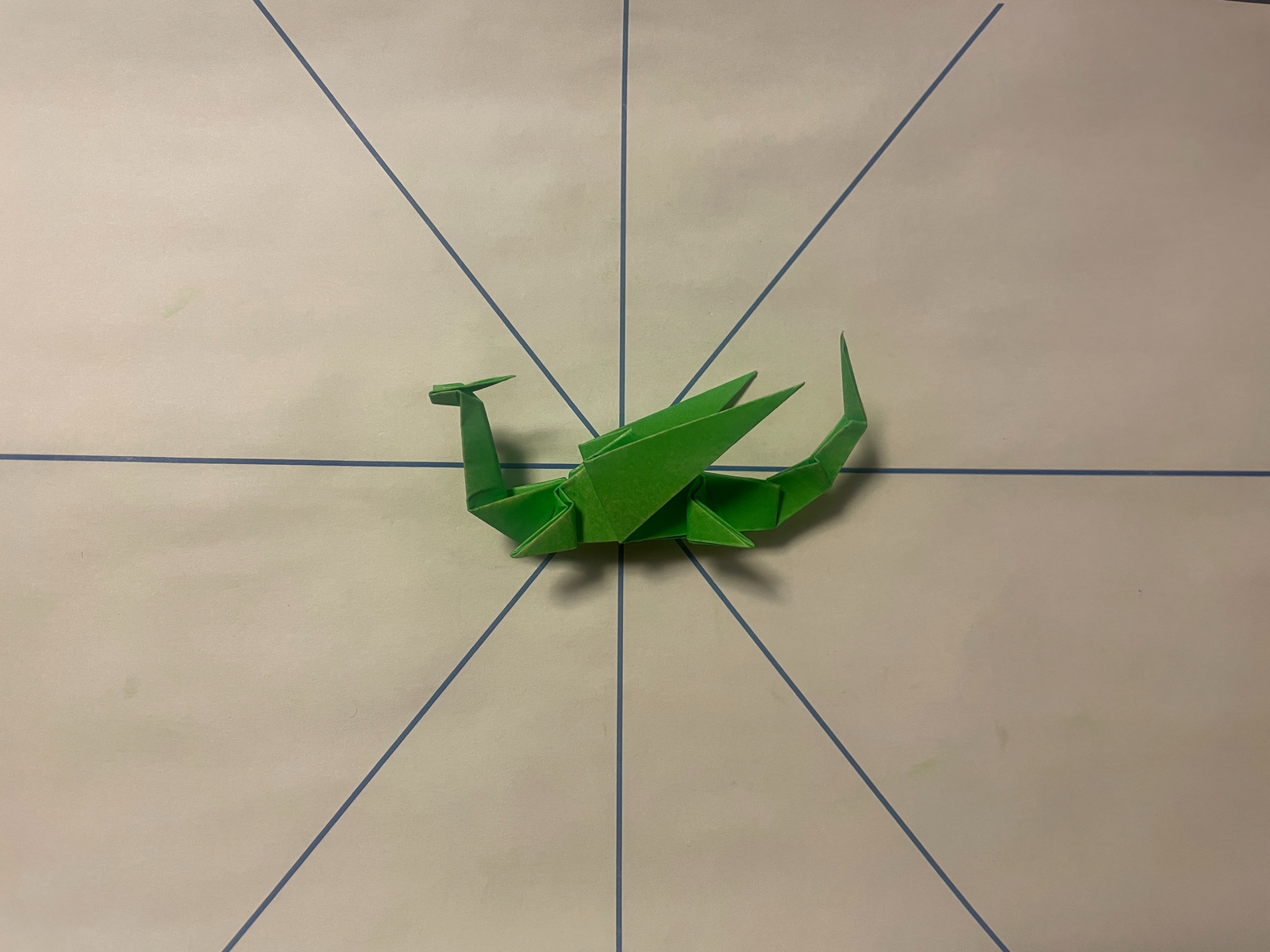}    
    \\
Cat
& Long-Horned Samurai Helmet 
& Crane 
& Dragon\\
    (a) Beginner
&        (b) Elementary
&            (c) Intermediate
&                (d) Advanced\\
        \end{tabular}
    \caption{
    Example designs for each difficulty level.
    }
    \label{fig:difficulty}
\end{figure}

\paragraph{Difficulty annotation.}
Each origami design in the dataset was annotated with a difficulty level.
In this study, difficulty was defined and categorized into five levels (\cref{tab:difficulty}) based on the number of procedural steps and the complexity of folding operations.
The annotations were performed independently by three annotators.
\cref{tab:difficulty} also shows the distribution of the final difficulty levels.
Final difficulty levels were determined by majority vote, i.e., agreement by at least two annotators; no design was assigned three different levels.
Inter-annotator agreement was Krippendorff's $\alpha = 0.88$, with a full-agreement rate of 80.2\%.
Examples of each difficulty level are shown in~\cref{fig:difficulty}.

\section{Benchmark simulation}
\label{sec:Benchmark-Simulation}

Simulator-based experiments were conducted for Tasks 1 and 2 using the proposed benchmark dataset.
As described in~\cref{sec:Task-Definition}, these experiments were performed to evaluate whether a multimodal LLM controller can understand origami state transitions from visual and textual inputs and ground them into structured simulator commands.\footnote{These experiments do not evaluate physical robot manipulation, low-level motor control, or full recovery from physically deviated paper states.}

\subsection{Common settings}

\paragraph{Agents.}
Three multimodal LLM controllers were used for both tasks, i.e., GPT 5.6 Sol, Gemma 4 31B, and Gemma 3 12B IT, listed in decreasing order of expected performance.
The primary goal is to provide an initial demonstration of how the dataset can be used to evaluate procedural state-transition reasoning in an origami simulator.
An additional goal is to investigate whether the benchmark can distinguish controllers with different capability levels.
For a controlled comparison, we used the same episodes, prompts, command interface, and attempt budget across all three controllers.
The model names, prompts, command schema, and representative examples are given in~\cref{appendix:benchmark-simulation-details}.

\paragraph{Episodes.}
This evaluation utilized all 348 reference origami procedures in the benchmark dataset.
These procedures are guaranteed to be foldable by a person following the written instructions (\cref{subsec:dataset_construction_procedure}); thus this setup was sufficient to evaluate whether the LLM controllers could also complete the same transitions.
For each procedure, we randomly selected a start state and set the target state to be no greater than 10 transitions later, or at the final state if fewer than 10 transitions remained. Note that the start state may already be partially folded rather than a flat sheet, and it is initialized in the simulator as the corresponding state in the dataset (\cref{appendix:episode-initialization}).
Furthermore, the same 348 procedures and the same start--target pairs were used for both Tasks 1 and 2.
However, the transitions actually attempted may differ between these tasks because an episode is terminated when any transition cannot be completed within the determined attempt budget.
We limited the segment length because evaluating full procedures would make episode success too sparse for an initial study.

\paragraph{Simulator interface.}
Unless otherwise stated, the conversation was reset at the beginning of each episode.
An episode was terminated when all target transitions from the sampled start state to the target state were judged successful, or when any transition could not be completed within the provided attempt budget.
Each transition permitted at most three attempts.
The simulator commands were expressed in JSON format and included various operations, e.g., \texttt{fold}, \texttt{make\_crease}, and \texttt{rotate\_3d}. Note that a \texttt{fold} could also tuck the folded flap into a pocket or layer gap by specifying an optional insertion target.
Although the simulator interface includes a \texttt{finish} command for future extensions, it was not used in the present experiments because trial termination was determined by evaluator judgment.

\paragraph{Retry-based correction.}
The current simulator does not reliably support corrective manipulation from arbitrary deviated folded states; thus, we evaluated retry-based correction as a proxy for recovery-related behavior.
As discussed previously, Tasks 1 and 2 used fixed feedback after failed attempts and free-form feedback from a human instructor, respectively.

\paragraph{Human evaluation.}
For both tasks, human evaluators judged the correctness of the LLM-generated command(s).
To assess the reliability of this judgment, a second evaluator, blind to the original verdicts, independently re-judged the commands generated for the transitions of the same 50 origami designs in both tasks, yielding 86.3\% agreement with the original judgments.

\paragraph{Evaluation metrics.}
As discussed in~\cref{subsec:evaluation-philosophy}, both tasks were evaluated using five metrics.
The primary metric was the Episode-SR (successful episodes / evaluated episodes).
As transition-level diagnostics, we also calculated and reported the Trans-SR (successful target transitions / all target transitions), First-Att-SR (first-attempt successes / attempted target transitions), and Avg Att/Trans (total command attempts / attempted target transitions), and we measured the retry-based correction with Retry-SR (successful retry corrections / failed first attempts).

\subsection{Task 1}

\paragraph{Task setting.}
In Task 1, the LLMs received all original images from the start state to the target state, together with the textual instructions for the corresponding transitions.
At each transition, the LLM output the simulator command(s) intended to move the current state to the next reference state.

\paragraph{Results.}
The results for Task 1 are summarized in~\cref{tab:benchmark_simulation_results}.
Episode-SR was low overall. Here, Gemma 4 31B and Gemma 3 12B IT completed only a small fraction of episodes, while GPT 5.6 Sol (the best-performing controller) obtained an Episode-SR of only 22.7\%, indicating that this task is challenging for current LLMs.
At the transition level, GPT 5.6 Sol achieved a Trans-SR of 34.1\%, followed by Gemma 4 31B (6.4\%) and Gemma 3 12B IT (0.8\%).
Furthermore, all pairwise differences in Trans-SR were significant, suggesting that the benchmark can be used to distinguish performance differences among controllers.
For GPT 5.6 Sol and Gemma 4 31B, Trans-SR was higher than Episode-SR, which indicates that completing some individual transitions did not ensure completion of entire procedure segments.
Moreover, GPT 5.6 Sol obtained the highest First-Att-SR (61.3\%), suggesting that it could frequently translate the image-text procedure into a correct initial command.
In addition, Retry-SR remained at only 38.2\% even for GPT 5.6 Sol, suggesting that fixed feedback provides limited support for revising failed commands.
Overall, these results show that the evaluated LLMs can attempt multimodal origami state transitions in a controlled simulator but still struggle to complete procedure segments reliably.

\begin{table}[t]    
\centering
\small
\setlength{\tabcolsep}{5pt}
\caption{
Benchmark simulation results for Tasks 1 and 2 across three LLM controllers.
SR denotes success rate and Att represents attempt(s).
Parenthesized values indicate numerator/denominator counts.
Lower Avg Att/Trans is better; higher success rates are better.
$^{*}$ indicates a significant improvement over Gemma 4 31B, and $^{\dagger}$ indicates a
significant improvement over Gemma 3 12B IT, based on paired bootstrap tests resampling origami designs (10,000 resamples; significant when the 95\% interval of the difference excludes zero).
}
\label{tab:benchmark_simulation_results}
\begin{tabular}{l rrrrr}
\toprule
Model & Episode-SR & Trans-SR & First-Att-SR & Avg Att/Trans & Retry-SR \\
\midrule
\multicolumn{6}{l}{\textit{Task 1}} \\
GPT 5.6 Sol    & $^{*\dagger}$22.7\% & $^{*\dagger}$34.1\% & $^{*\dagger}$61.3\% & $^{*\dagger}$1.68 & $^{*\dagger}$38.2\% \\
\   & (79/348) &  (855/2505) & (689/1124) & (1888/1124) & (166/435) \\
Gemma 4 31B    & $^{\dagger}$3.7\% & $^{\dagger}$6.4\% & $^{\dagger}$26.3\% & $^{\dagger}$2.44 & $^{\dagger}$8.2\% \\
\   & (13/348) & (160/2505) & (130/495) & (1208/495) & (\; 30/365) \\
Gemma 3 12B IT & 1.7\% & 0.8\% & 2.8\% & 2.93 & 2.6\% \\
\   & (\; 6/348) & (\; 19/2505) & (\; 10/361) & (1056/361) &  (\; \; 9/351) \\
\midrule
\multicolumn{6}{l}{\textit{Task 2}} \\
GPT 5.6 Sol    & $^{*\dagger}$25.6\% & $^{*\dagger}$40.3\% & $^{*\dagger}$58.5\% & $^{*\dagger}$1.69 & $^{*\dagger}$50.9\% \\
\  & (89/348) & (1010/2505) & (742/1269) & (2139/1269) & (268/527) \\
Gemma 4 31B    & 3.4\% & $^{\dagger}$8.5\% & $^{\dagger}$28.6\% & $^{\dagger}$2.36 & $^{\dagger}$14.1\% \\
\  & (12/348) & (212/2505) & (157/548) & (1296/548) & (\; 55/391) \\
Gemma 3 12B IT & 2.9\%  & 3.0\%  & 14.3\%  & 2.69 & 4.8\% \\
\  &  (10/348) & (\; 76/2505) & (\; 59/414) & (1112/414) & (\; 17/355) \\
\bottomrule
\end{tabular}
\end{table}

\subsection{Task 2}

\paragraph{Task setting.}
Task 2 used the same dataset, sampling procedure, LLMs, simulator interface, attempt budget, and human judging criteria as Task 1.
However, unlike Task 1, the LLMs did not directly receive the reference image sequence, textual procedure, or target state.
Instead, a human instructor observed the reference procedure and the current state image, and then provided step-by-step guidance in natural language.
Here, the instructor was permitted to clarify the relevant instruction and provide corrective feedback, but was not permitted to directly issue simulator commands or reveal the full future sequence.  
In these experiments, the instructor guidance and agent questions were required to use natural geometric descriptions rather than refer to the dot stickers or guide lines.

\paragraph{Results.}
\cref{tab:benchmark_simulation_results} summarizes the results for Task 2.
Both Tasks 1 and 2 used the same procedures and start--target pairs; thus, the comparison controls for episode selection.
As in Task 1, Episode-SR was low overall. Here, Gemma 4 31B and Gemma 3 12B IT completed only a small fraction of episodes, while GPT 5.6 Sol (the best-performing controller) reached 25.6\%.
At the transition level, GPT 5.6 Sol obtained a Trans-SR of 40.3\%, followed by Gemma 4 31B (8.5\%) and Gemma 3 12B IT (3.0\%).
As in Task 1, all pairwise differences in Trans-SR were significant.
These results show that the benchmark distinguishes controller performance consistently in both the autonomous and human-guided settings.
Compared with Task 1, Trans-SR and Retry-SR were higher in Task 2 for all three controllers.
This suggests that, even without direct access to the reference procedure, instructor guidance may help the controllers complete individual transitions and revise failed commands, although its effect on completing entire procedure segments was limited.

\subsection{Human-guided interaction analysis}
\label{sec:human-guided-interaction-analysis}

\paragraph{Analysis motivation.}
As shown in~\cref{tab:benchmark_simulation_results}, both Trans-SR and Retry-SR were higher in Task 2 than in Task 1 for all three controllers.
Thus, we analyzed two interactions specific to Task 2, i.e., clarification questions and retry-based correction, to better understand the factors contributing to the observed improvement.

\begin{table}[t]
\centering
\small
\tabcolsep 4pt
\caption{
Success rates on identical transitions in Tasks 1 and 2, split by whether the agent asked a clarification question in Task 2.
Only transitions attempted in both tasks are included.
Parenthesized values indicate numerator/denominator counts.
$^{*}$ indicates a significant difference from Task 1, based on paired bootstrap tests resampling origami designs (10,000 resamples; significant when the 95\% interval of the difference excludes zero).
}
\label{tab:question_transition_comparison}
\begin{tabular}{lrrrr}
\toprule
& \multicolumn{2}{c}{Without clarification questions} & \multicolumn{2}{c}{With clarification questions} \\
\cmidrule(lr){2-3} \cmidrule(lr){4-5}
Model & Task 1 & Task 2 & Task 1 & Task 2 \\
\midrule
GPT 5.6 Sol    & 81.7\% (642/\; 786) & $^{*}$86.9\% (683/\; 786) & 48.8\% (78/160) & 40.6\% (65/160) \\
Gemma 4 31B    & 37.6\% (132/\; 351) & $^{*}$45.3\% (159/\; 351) & 13.5\% (15/111) & 9.0\% (10/111) \\
Gemma 3 12B IT & 5.7\% (\; 15/\; 263) & $^{*}$24.7\% (\; 65/\; 263) & 4.2\% (\; 4/\; 95) & 5.3\% (\; 5/\; 95) \\
\midrule
All            & 56.4\% (789/1400) & $^{*}$64.8\% (907/1400) & 26.5\% (97/366) & $^{*}$21.9\% (80/366) \\
\bottomrule
\end{tabular}
\end{table}

\paragraph{Clarification questions.}
In Task 2, agents can ask clarification questions after receiving the instructor's guidance.
\cref{tab:question_transition_comparison} compares Task 1 and Task 2 success rates on transitions attempted in both tasks, grouped by whether a clarification question was asked in Task 2.
Clarification questions occurred in 17.7\%, 23.7\%, and 27.8\% of attempted Task 2 transitions for GPT 5.6 Sol, Gemma 4 31B, and Gemma 3 12B IT, respectively.
For transitions with clarification questions, the Task 1 success rate was 26.5\%, compared with 56.4\% for those without, suggesting that questions tend to arise on difficult transitions.
Despite receiving answers, the Task 2 success rate on these transitions was only 21.9\%, significantly lower than in Task 1.
In contrast, Task 2 significantly outperformed Task 1 on transitions without clarification questions.
These findings suggest that instructor guidance alone may be insufficient for difficult transitions, whereas direct access to reference images and the target state, as in Task 1, may provide more useful information.

\paragraph{Retry-based correction.}
In the current simulator, retry-based correction serves as a proxy for recovery-related behavior.
As shown in~\cref{tab:benchmark_simulation_results}, Retry-SR was higher in Task 2 than in Task 1 for all three controllers (GPT 5.6 Sol: 38.2\%$\rightarrow$50.9\%; Gemma 4 31B: 8.2\%$\rightarrow$14.1\%; Gemma 3 12B IT: 2.6\%$\rightarrow$4.8\%).
Both tasks used the same procedures and start--target pairs; thus, these results suggest that interactive feedback from a human instructor can help the LLM controllers revise failed commands more effectively than fixed feedback alone.
This result provides initial evidence supporting the value of the human-guided setting in the proposed benchmark.
However, it should be interpreted cautiously.
The current simulator does not evaluate full recovery from physically deviated paper states, and the performance of Task 2 may depend on instructor phrasing, clarification behavior, corrective-feedback style, and uneven instructor workload.
Additional discussion is provided in~\cref{appendix:human-guided-correction-analysis}.

\section{Limitations}
\label{sec:limitation}
Several limitations must be acknowledged.
First, the experiments conducted in this study involved simulation, and whether the simulated operations are feasible for physical robots was not assessed.
Thus, it is possible that the results do not generalize to real-world robotic execution.
Second, the simulator does not support full recovery from folded states with arbitrary deviations; therefore, retry-based correction only assesses whether the agent can revise a command in response to feedback while the current state image remains unchanged.
Third, the dataset is controlled and fiducial-assisted, with radial guide lines, two-sided colored paper, and occasional dot stickers; thus, the results may not directly transfer to unconstrained visual conditions.
Fourth, the evaluation was dependent on human judgment, and in Task 2, the instructor for each episode also served as the evaluator for that episode.
The Task 2 performance may also depend on instructor phrasing, clarification behavior, corrective-feedback style, and uneven workload distribution across instructors.
Finally, the experiments were limited to a single simulator, three multimodal LLM controllers, and segments of at most 10 transitions per procedure.

\section{Conclusion}

This paper has introduced origami as a real-image benchmark for procedural state-transition reasoning.
In this study, we constructed \textsc{OrigamiTrace}, a dataset of real step-by-step origami images with textual entries, difficulty annotations, and final-state plausibility validation, and we defined two tasks, i.e., autonomous state transitions from multimodal instructions and human-guided collaborative transitions.
In an initial simulator-based experimental evaluation, we tested three multimodal LLM controllers through an origami simulator, focusing on procedure understanding, action-interface grounding, and retry-based correction rather than on physical robot manipulation or full recovery from deviated fold states.
Future work will improve simulator support for complex folds and full recovery, strengthen transition-level dataset validation, and extend the proposed benchmark to additional models, simulators, physical robotic platforms, and VLA-based embodied agents.

\clearpage
\section*{Acknowledgments}

This work was supported
by JSPS KAKENHI Grant Numbers JP26H02511, JP25K21263, JST BOOST Grant Number JPMJBY24A1, and
JST Moonshot R\&D Grant Number JPMJMS2011-35 (fundamental research).

We thank all Baobab project managers, including Toshie Amemiya, Baochecker quality checkers, and all BaoExpert AI trainers who contributed to the construction of this dataset. 
We are particularly grateful to the participants and staff of the employment support facilities for individuals with disabilities for their contributions to image collection and caption annotation.

{\small
\bibliographystyle{plain}
\bibliography{main}
}

\clearpage
\appendix

\section{Additional motivation and positioning}
\label[appendix]{appendix:origami-motivation}

This appendix expands the motivation for using origami as a real-image procedural state-transition benchmark.
The main paper states the central claims compactly; here we explain why those claims are supported.

\subsection{Origami as a procedural state-transition task}
\label[appendix]{appendix:motivation-procedural-task}

Origami execution is naturally formulated as a sequence of state transitions.
At each step, an agent must recognize the current folded state, infer the intended next state, and choose an action that transforms the paper accordingly.
This differs from evaluating whether an agent can recognize a final shape or reason from an abstract crease pattern.
In actual folding, the current paper state includes geometry, orientation, visible and hidden layers, accumulated creases, and local reference points.
The meaning of an instruction often depends on this current state.

\subsection{Why final-state evaluation alone is insufficient}
\label[appendix]{appendix:motivation-final-state-insufficient}

Final-state evaluation can determine whether the folded origami appears complete, but it does not reveal where or why an agent failed.
Small local errors can persist across later steps: a misplaced crease may change layer ordering, hide a future reference point, or make a later insertion infeasible.
Therefore, a benchmark based only on final completion cannot diagnose whether an agent failed due to state recognition, layer selection, local action grounding, or accumulated errors.
This motivates our use of episode success rate together with transition-level metrics and retry-based correction.

\subsection{Why real intermediate folding states are necessary}
\label[appendix]{appendix:motivation-real-intermediate-states}

Many origami-inspired benchmarks evaluate spatial reasoning from crease patterns, diagrams, or puzzle-like images.
These representations are useful for studying geometric imagination, but they abstract away the visual states encountered during actual folding.
In real procedures, the agent must interpret the current folded state, including visible and hidden layers, partial occlusions, orientation changes, accumulated creases, and local reference points.
Our benchmark therefore uses real step-by-step images of folded paper states rather than only abstract representations.

\subsection{Why origami is useful among procedural manipulation tasks}
\label[appendix]{appendix:motivation-procedural-domain}

Origami shares important properties with procedural manipulation domains such as cooking, assembly, and repair.
It requires an agent to follow instructions, ground language in the current physical state, perform local operations, and correct mistakes.
However, origami has practical advantages for benchmarking.
A task can often be instantiated with a single sheet of paper, the material is inexpensive, the initial state is easy to reproduce, and the final folded shape provides a visually clear goal.
In our benchmark setting, we exclude tasks requiring scissors, adhesives, or multiple independent sheets, reducing dependence on specialized tool-use skills.
This makes origami easier to standardize than many real-world procedural tasks while retaining substantial perceptual, spatial, and sequential difficulty.

\subsection{Why human-guided execution is included}
\label[appendix]{appendix:motivation-human-guided-execution}

Real embodied agents may not always receive complete task specifications in a machine-readable form.
Instead, they may need to collaborate with humans who provide partial instructions, clarification, and corrective feedback.
Task 2 is intended to evaluate this instructor-mediated setting.
The agent does not directly receive the full reference image sequence, textual procedure, or target state, but must act based on current observations and information communicated through dialogue.
This setting tests whether an agent can use human-provided information to complete the same state-transition objective, rather than merely following a fully specified procedure.

\subsection{Why retry-based correction is evaluated}
\label[appendix]{appendix:motivation-retry-based-correction}

Full physical recovery would require the environment to preserve deviated paper states and allow corrective manipulation from those states.
This is important for future physical robot evaluations, but is not reliably supported by the current simulator.
Retry-based correction provides a simplified proxy: after a failed attempt, the simulator keeps the current state image unchanged and the agent retries the same transition after receiving feedback.
Although this does not measure full recovery from a physically deviated state, it evaluates whether the agent can revise its action decision after failure, which is an important component of robust procedural execution.

\subsection{Broader impacts}
\label[appendix]{appendix:broader-impacts}
This benchmark may support safer and more systematic evaluation of embodied AI systems before physical deployment, especially for fine-grained procedural tasks. Potential negative impacts include overinterpreting simulator-based results as physical robot capability and future misuse of improved embodied agents in unsafe physical settings. We mitigate these risks by clearly limiting our claims to benchmark formulation, dataset construction, and simulator-based initial evaluation.

\section{Positioning relative to origami-inspired benchmarks}
\label[appendix]{appendix:origami-positioning}

Origami has appeared in prior research in several forms, including spatial reasoning benchmarks, paper-folding puzzles, robotic origami, and deformable manipulation.
The main distinction of our work lies in the use of origami and, in particular, in how origami is used.

\subsection{Crease-pattern and diagram-based reasoning}
\label[appendix]{appendix:positioning-crease-diagram}

Crease patterns and diagrams provide abstract geometric descriptions of origami.
They are useful for evaluating whether a model can infer folded structures, reason about symmetry, or imagine 3D shapes from 2D information.
However, they do not directly evaluate how an agent should act from a real intermediate folded state.
In actual folding, the agent must determine which layers are accessible, which references are visible, and how a local fold changes the current physical configuration.

\subsection{Puzzle-like paper folding tasks}
\label[appendix]{appendix:positioning-puzzle-tasks}

Paper-folding puzzles often test mental rotation, spatial transformation, or 2D-to-3D reasoning.
These tasks are valuable for evaluating spatial cognition, but they usually do not require executing a procedure over real intermediate states.
They also do not typically evaluate correction behavior after an incorrect action.

\subsection{Robotic origami}
\label[appendix]{appendix:positioning-robotic-origami}

Robotic origami studies address important physical challenges such as grasping, folding mechanics, paper deformation, and controller development.
These studies often focus on specific robot mechanisms or particular folding strategies.
In contrast, our benchmark abstracts over the embodiment and defines task formats and metrics for comparing agents in simulation and future physical settings.
The goal is not to propose a specific origami robot, but to use origami as a domain for evaluating procedural state-transition reasoning.

\subsection{Our setting}
\label[appendix]{appendix:positioning-our-setting}

Our benchmark uses real step-by-step images of actual folded paper states with natural language instructions.
The agent must reason from the current observed state to the next intended state, rather than only solving an abstract spatial reasoning problem.
This makes the benchmark complementary to prior origami-inspired work: it preserves the geometric richness of origami while shifting the focus toward procedural execution, correction behavior, and future embodied evaluation.

We also note contemporaneous work, OrigamiBench~\citep{agarwal2026can}, whose revised paper was released in October 2026.
OrigamiBench uses a high-level folding action language and evaluates transition reasoning and program grounding through multiple-choice next-state prediction.
We view OrigamiBench as work developed in parallel with ours, and as complementary rather than redundant. 
Its focus is on understanding programmatic folding actions and their effects in rendered paper states, while our benchmark focuses on executing procedural state transitions grounded in real step-by-step images and human-written textual instructions.
In addition, our Task 2 studies instructor-mediated execution, where the agent acts from current observations and human guidance rather than from direct access to the full reference procedure.

\section{Detailed comparison of Tasks 1 and 2}
\label[appendix]{appendix:task-comparison-detail}
\cref{tab:task_comparison_detail} shows the detailed list of comparisons between Tasks 1 and 2.

\begin{table}[t]
\centering
\small
\tabcolsep 3pt
\caption{
Detailed comparison of the two benchmark tasks.
RIS: reference image sequence; RTI: reference textual instructions.
}
\label{tab:task_comparison_detail}
\begin{tabular}{p{0.22\linewidth}p{0.36\linewidth}p{0.36\linewidth}}
\toprule
Aspect 
& Task 1: Autonomous Origami State Transition 
& Task 2: Human-Guided Collaborative Origami State Transition \\
\midrule

Question 
& Can the agent complete the procedure segment by directly using the reference procedure? 
& Can the agent complete the same procedure segment through human guidance? \\

Core setting 
& Direct multimodal procedure following. 
& Instructor-mediated state-transition reasoning. \\

Information flow 
& Dataset $\rightarrow$ Agent. 
& Dataset $\rightarrow$ Human instructor $\rightarrow$ Agent. \\

Agent access 
& Given RIS, RTI, and target state. 
& No direct access to RIS, RTI, or target state. \\

Instructor access 
& Not used. 
& The instructor has access to RIS, RTI, and the target state. \\

Instruction form 
& Image-text procedure. 
& Step-by-step natural language guidance. \\

Visual reference use 
& RIS is shown to the agent. 
& RIS is hidden from the agent and observed by the instructor. \\

Reference text use 
& RTI is read directly by the agent. 
& The instructor may read, paraphrase, or locally clarify RTI. \\

Interaction 
& No instructor dialogue. 
& Allowed through instructor guidance, clarification, and feedback. \\

Clarification 
& Not allowed. 
& The agent may ask clarification questions, and the instructor may answer them. \\

Human intervention 
& No instructor guidance or physical manipulation. 
& The instructor may communicate with the agent but may not manipulate the paper or directly operate the environment. \\

Action command 
& Generated by the agent. 
& Generated by the agent; the instructor may not directly specify simulator commands. \\

Correction setting 
& Retry-based correction with fixed feedback. 
& Retry-based correction with free-form instructor feedback. \\

Reference-sequence disclosure 
& The relevant reference sequence is directly available to the agent. 
& The instructor may not disclose the full future reference sequence in advance. \\

Primary metric 
& Episode success rate. 
& Episode success rate. \\

Transition-level metrics 
& Transition success rate, first-attempt success rate, and average attempts per transition. 
& Same as Task 1. \\

    Correction metric 
& Retry-correction success rate after failed first attempts. 
& Retry-correction success rate after failed first attempts. \\

Interaction statistics 
& Not applicable. 
& Instructor turns, clarification questions, feedback turns, and instruction length are logged. \\

Benchmark role 
& Tests direct procedure understanding, visual-language grounding, and fine-grained state-transition reasoning. 
& Tests human-guided state-transition reasoning, collaborative grounding, and correction through dialogue. \\

Current simulation setting 
& The LLM generates structured simulator commands from reference images and textual instructions. 
& The LLM generates structured simulator commands from the current simulator observation and instructor guidance. \\

Recovery interpretation 
& Measures retry-based correction, not full recovery from a physically deviated paper state. 
& Measures human-guided retry-based correction, not full recovery from a physically deviated paper state. \\

\bottomrule
\end{tabular}
\end{table}

\section{Details of data construction}
\label[appendix]{appendix:dataset-construction}

This appendix provides additional details on dataset access, reference-asset handling,
worker-facing instructions, contracted data creation, quality control, and final-state
plausibility evaluation.

\subsection{Data access, license, and reference assets}
\label[appendix]{appendix:origami-license}

The dataset is available at:
\url{https://doi.org/10.7910/DVN/V6T9LX}.
The released dataset is licensed under CC-BY-SA 4.0.

To collect candidate origami designs, we used publicly accessible origami-instruction
websites for early childhood education and recreational use as procedural references.
We used design names and procedural information, but did not redistribute original website
images, diagrams, videos, page layouts, or textual instructions.
The released dataset consists of newly created assets: trained workers manually folded
the designs, captured new step-by-step images, and wrote new textual entries according to
the worker-facing specification described in~\cref{appendix:worker-facing-specification}.

\subsection{Worker-facing specification and data-creation workflow}
\label[appendix]{appendix:worker-facing-specification}

\begin{figure}[t]
    \centering
    \includegraphics[width=0.9\linewidth]{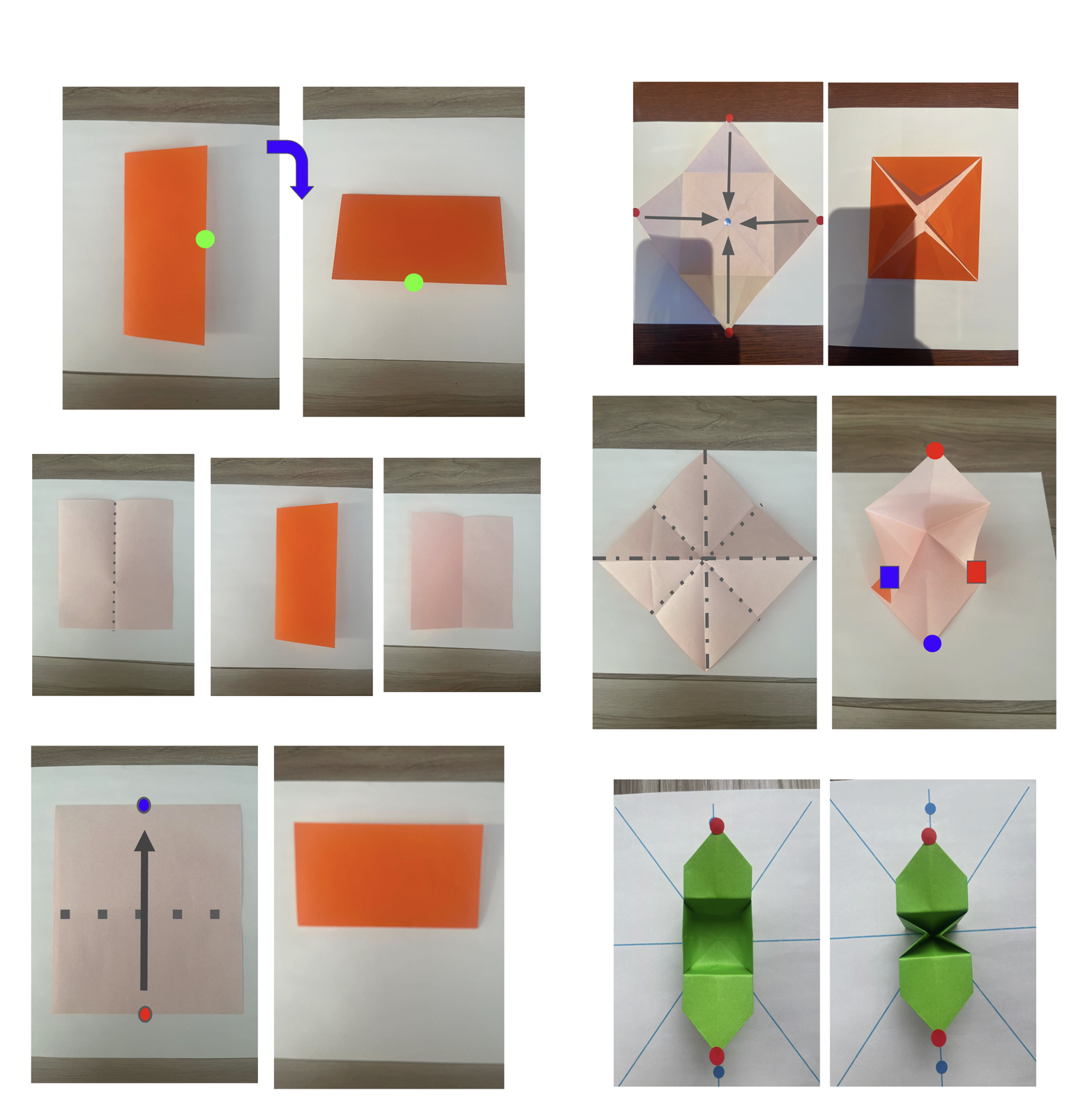}
    \caption{
    Representative excerpts from the worker-facing data-creation specification.
    The specification described the image--text data format, standardized folding terminology,
    fiducial sticker conventions, scope restrictions, and file-naming rules.
    }
    \label{fig:appendix-worker-facing-spec}
\end{figure}

The dataset was created through manual data-production work based on a 
worker-facing specification document.
The work did not use a dedicated annotation interface.
Instead, workers manually folded origami designs, captured step-by-step images, and
created textual entries according to the specification.
Representative excerpts from the specification are shown in~\cref{fig:appendix-worker-facing-spec}.

The specification was intended for future evaluation of robotic or embodied agents.
It required each data item to consist of an origami image and a textual entry.
Images were not to contain explanatory text, arrows, dashed lines, or other instructional
overlays; information needed to identify the intended operation was expressed in the
textual entry.
This choice keeps the visual input close to an observed paper state while making the
procedural instruction explicit in text.

The specification standardized folding terminology before captioning.
It defined basic operations such as \textit{valley fold}, \textit{mountain fold},
\textit{make a crease}, \textit{clockwise} and \textit{counterclockwise rotation}, \textit{standing the origami
upright}, and \textit{turning it over}.
It also defined frequently used compound operations such as \textit{square fold},
\textit{triangle fold}, \textit{inside reverse fold}, \textit{outside reverse fold},
and \textit{insert}.
This terminology was used to reduce variation in textual entries across workers and designs.

The specification allowed workers to use colored dot stickers to make local directions
and reference positions explicit.
Red and blue stickers indicated source and target positions for folds when needed.
For insertion operations, a yellow sticker could be used as an intermediate passing point.
These fiducial cues were intended to make the relevant local operation visible while
preserving the real folded-paper state.

The specification excluded designs or steps unsuitable for the intended embodied-agent
setting, including designs requiring scissors, adhesives, or multiple independent sheets
of paper.
Procedures requiring simultaneous actions judged incompatible with the intended
single-agent paper-manipulation setting, such as pinching and holding multiple parts at
once, were also treated as outside the target scope.
The specification further defined file-naming rules so that procedural images could be
indexed by design name, optional three-digit suffix, and step number.

The dataset was constructed in collaboration with Baobab, Inc., through a contracted workflow involving project managers, Baochecker quality checkers, and BaoExpert AI trainers. 
The image-collection stage involved 27 individuals, including three project managers, 13 quality checkers, and 11 BaoExpert AI trainers from three employment support facilities for individuals with disabilities (including three support staff members). 
The captioning stage involved 25 individuals, including three project managers, 10 quality checkers, and 12 BaoExpert AI trainers from two such facilities (including three support staff members). 
All BaoExpert AI trainers, including support staff, were required to complete Baobab Academy, Baobab's labeling training program, before participating in the project.

The correction and evaluation stage involved 24 individuals, including two project managers, 13 quality checkers, and nine BaoExpert AI trainers from two employment support facilities (including three support staff members). 
The quality checkers, including project-management members and checker leaders, supported the multi-stage quality-control workflow described in~\cref{appendix:transition-quality-control}.

Image collection was conducted over approximately one month, with a corresponding checking period of approximately five weeks. Captioning was conducted over 13 days, with a corresponding checking period of 20 days.

This workflow supports the dataset as a curated collection of real origami procedure images and textual entries. As detailed in~\cref{appendix:transition-quality-control}, transition-level image-text consistency was checked during data creation through this multi-stage workflow. 
We distinguish this process-level quality assurance from a separate large-scale post-hoc statistical validation of every transition, which is left for future work.

\subsection{Transition-level quality control}
\label[appendix]{appendix:transition-quality-control}

Transition-level consistency was checked during the data-creation workflow. Each procedure was produced under a worker-facing specification that standardized the image-text format, folding terminology, fiducial-sticker conventions, scope restrictions, and file-naming rules. Workers first reproduced the folding process and captured step-by-step images, and textual entries were then created to describe the operation from each state to the next state. During checking, reviewers verified that the textual entry followed the predefined terminology, referred to visible fiducial cues consistently when needed, and described the intended transition between adjacent paper states.

The workflow involved multiple roles, including project managers, image-collection workers, captioning workers, and dedicated checkers. The checker group included project-management members and checker leaders, forming a multi-stage quality-control process. Image collection and captioning were each followed by separate checking periods. Therefore, transition-level image-text consistency was not left to a single annotator; it was controlled through predefined specifications, fiducial-assisted capture, trained workers, and multi-stage human checking.

After data creation, we applied the quality checking described in \emph{Quality Checking} in~\cref{subsec:dataset_construction_procedure} to every transition of every design.
The specific check items are listed in~\cref{tab:quality_check_items}.
For each transition, checkers compared the images of the two adjacent paper states with the corresponding textual entry and, for each of the three perspectives, judged whether the transition had no issues, required an image correction, or required a text correction.
To judge foldability, checkers did not rely on visual comparison alone: in every round, they folded a sheet of origami paper themselves from the beginning of the folding procedure along with the review, and for each transition confirmed that folding from the state shown in the earlier image by following the textual entry yielded the state shown in the next image.
Transitions judged to require correction were corrected in the image or the textual entry following the same procedure as in the original creation, and a different checker from the previous round re-checked them with the same check items.
This correction-and-recheck cycle was repeated until every transition was judged to have no issues in all three perspectives.
Consequently, every transition in the dataset has been confirmed to be foldable by a person from the given state to the next by following the written instruction.

\begin{table}[t]
\centering
\small
\tabcolsep 2pt
\caption{
Check items used in quality checking.
For each transition, checkers judged each of the three perspectives as having no issues, requiring an image correction, or requiring a text correction (both corrections could be required).
}
\label{tab:quality_check_items}
\begin{tabular}{p{0.17\linewidth}p{0.25\linewidth}p{0.55\linewidth}}
\toprule
Perspective & Item & What is checked \\
\midrule
Foldability
& Foldability & Whether a person can fold from the given state to the next by following the textual entry. \\
& Continuity & Whether no procedural step is skipped or duplicated between the adjacent paper states. \\
& Consistency & Whether the folded layers, rotations, and turn-overs are consistent with the images. \\
& Visibility of references & Whether fiducial cues such as dot stickers are visible in the images. \\
& Agreement with the change & Whether the operation in the textual entry matches the change between the adjacent paper states. \\
& Initial state & Whether the image before the transition shows the state before the operation begins (not a state in the middle of folding). \\
\midrule
Specification conformance
& Consistency with the specification & Whether the entry is consistent with the worker-facing specification (e.g., predefined terminology and references to fiducial cues). \\
\midrule
Text quality
& Wording & Whether there are no typos or inconsistent wording. \\
& Clarity & Whether the folding operation is not described ambiguously. \\
\bottomrule
\end{tabular}
\end{table}

In this quality checking, each transition was judged by one checker in each round.
A large-scale statistical re-annotation in which multiple raters independently judge every transition and formal inter-rater agreement is computed remains an important direction for future work.

Workforce information is reported in aggregate without identifying individual workers.
The data-creation tasks involved non-sensitive origami images and textual entries and did not require collection of personal or sensitive information from workers.

\subsection{Final-state plausibility evaluation}
\label[appendix]{appendix:dataset-quality-evaluation}

\begin{figure}[t]
    \centering
    \includegraphics[width=1.0\linewidth]{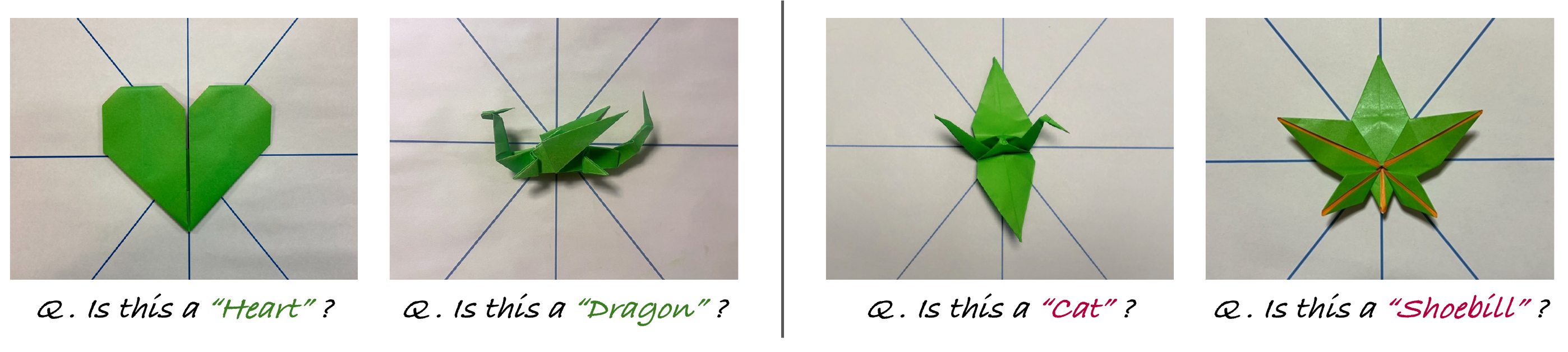}
    \caption{
    Evaluation of final-state plausibility using true labels and false labels.
    Crowdworkers answered \textit{yes} or \textit{no} to indicate whether the given image
    appeared to correspond to the specified label.
    }
    \label{fig:crowdsourcing}
\end{figure}

This section provides further details of the final-state plausibility evaluation described in~\cref{subsec:dataset_construction_procedure}, whose two settings are illustrated in~\cref{fig:crowdsourcing}.
Crowdworkers were recruited via Amazon Mechanical Turk\footnote{\url{https://www.mturk.com/}} for English speakers and Yahoo! Crowdsourcing\footnote{\url{https://crowdsourcing.yahoo.co.jp/}} for Japanese speakers.
English-speaking crowdworkers were required to reside in the United States, United Kingdom, or Canada, have a HIT approval rate of at least 98\%, and have completed more than 5,000 HITs.
Japanese-speaking crowdworkers were required to be native speakers with experience living in Japan.
All crowdworkers were required to demonstrate basic comprehension of the task instructions.
The evaluation used the standard interfaces provided by the corresponding platforms.

The expected response time was approximately 5--10 seconds per item.
Compensation was set at USD 0.50 per task batch of 11 items for English-speaking crowdworkers, corresponding to approximately USD 15 per hour, and JPY 21 per task batch for Japanese-speaking crowdworkers, based on the Tokyo minimum wage of JPY 1,226 per hour.
For quality control, one hidden golden task was inserted per task batch, and responses from crowdworkers who failed it were excluded.
Each item was ultimately evaluated by six crowdworkers: three English speakers and three Japanese speakers.

\begin{table}[t]
    \centering
    \small
    \caption{Label-decision accuracy for completed origami images.}
    \label{tab:quality_evaluation}
    \begin{tabular}{lccc}
        \toprule
        Settings
        & English speakers
        & Japanese speakers
        & All crowdworkers \\
        \midrule
        True-label evaluation
        & 71.3\% & 80.1\% & 75.7\% \\
        False-label evaluation
        & 88.1\% & 95.3\% & 91.7\% \\
        \midrule
        Overall
        & 79.7\% & 87.7\% & 83.7\% \\
        \bottomrule
    \end{tabular}
\end{table}

\cref{tab:quality_evaluation} breaks down the overall label-decision accuracy reported in the main text by setting and crowdworker group.
Accuracy was higher in the false-label setting than in the true-label setting, indicating that rejecting incorrect labels was easier than confirming some correct labels from a single final image.
We interpret this evaluation as a coarse validation of final-state plausibility rather than a guarantee of transition-level annotation correctness.

\paragraph{Ethics and participant risk.}
The data creation, final-state plausibility evaluation, and human-guided simulation
experiments involved external contractors, crowdsourcing workers, or experimenters
performing annotation/evaluation work on non-sensitive origami images and LLM-generated
commands.
The tasks did not involve deception, physical intervention, collection of personal or
sensitive information, or evaluation of worker attributes beyond platform-managed
eligibility criteria.
Workers were compensated through the corresponding platforms where applicable, and
the task procedures, worker eligibility criteria, compensation, and quality-control
procedures are described above.
No formal IRB approval or equivalent institutional review was obtained.

\section{Benchmark evaluation protocols}
\label[appendix]{appendix:benchmark-evaluation-protocols}

This appendix describes the actual evaluation protocols used for Tasks 1 and 2 in the
simulator-based experiments.

\paragraph{Experimental workforce.} 
The simulator-based experiments involved 21 individuals, including two project managers, 13 quality checkers (Baochecker), and six BaoExpert AI trainers from one employment support facility for individuals with disabilities (including one support staff member). 
These counts represent unique individuals involved in the experimental work.

\subsection{Task 1 protocol}
\label[appendix]{appendix:detailed-operation-for-task1}

Task 1 evaluates autonomous state transitions from multimodal reference information.
At the beginning of each episode, the LLM receives the reference image sequence from
the sampled start state to the target state, together with the corresponding textual
entries in CSV format.
For the initial input files, the LLM is instructed to output exactly \texttt{OK}.
For each subsequent step-specific request, the LLM receives the current reference image
and the corresponding textual instruction, and outputs exactly one non-empty code block
containing only the simulator command or commands for that step.

\begin{figure}[t]
    \centering
    \includegraphics[width=0.5\linewidth]{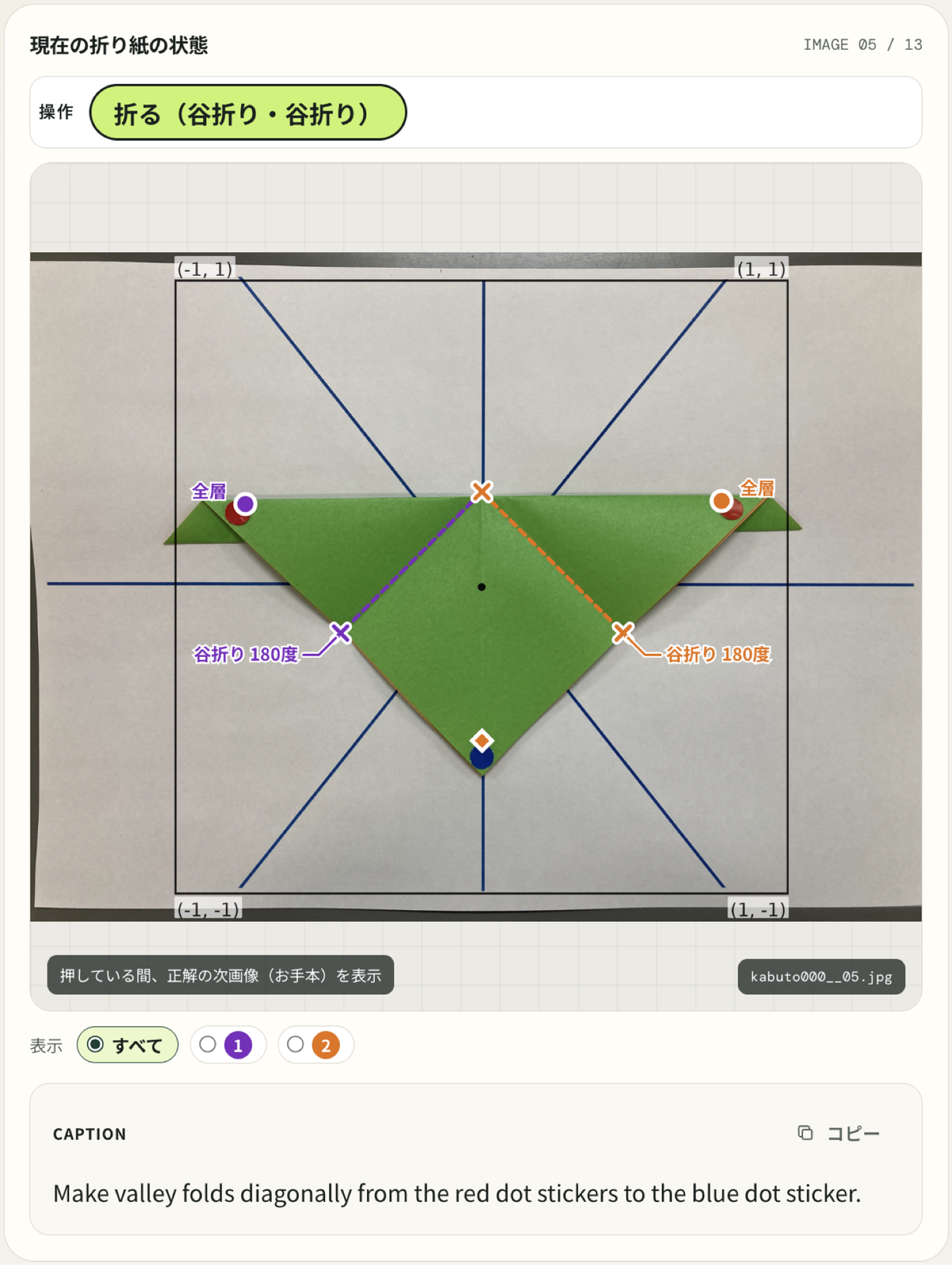}
    \caption{
    The human evaluation interface used in Tasks 1 and 2. The example shows a command for kabuto000
    (step 5) with two simultaneous valley folds, each bringing a marked source point
    (circle) to a shared target point (diamond); the crease line for each fold is marked
    by a pair of \texttt{×} symbols. A human evaluator judges whether the command is
    correct for reaching the reference next state.
    }
    \label{fig:task1-judging-interface}
\end{figure}

At each transition, the LLM outputs simulator command(s) intended to move the current
state to the next reference state.
In both tasks, one command attempt is defined as a single agent response containing one or more simulator commands.
Multiple commands may be included in the same attempt only if they do not interfere with one another, such that their execution order does not affect the resulting state.
A command attempt is counted as successful only if a human evaluator judges the generated command(s) to be correct for reaching the reference next state.
All other cases are counted as failed attempts, including invalid commands, simulator rejections, unsupported or conditionally unsupported transitions.
\cref{fig:task1-judging-interface} shows this evaluation interface.

When a failed attempt occurs, the LLM receives one of two fixed feedback messages:
\texttt{The command is invalid. Please regenerate it.} or \texttt{The fold is incorrect. Please regenerate it.}
Detailed simulator error messages, diagnostic codes, and internal error types are not
returned to the LLM.
If an attempt is judged unsuccessful by a human evaluator, the simulator keeps
the current state image unchanged, and the LLM retries the same transition.

Each transition allows at most three command attempts.
If the transition cannot be completed within this attempt budget, the episode terminates
and is marked as failed.
An episode is counted as successful only if all target transitions from the sampled start
state to the target state are completed within the attempt budget.
Each sampled episode is evaluated once.
The procedures, start states, target states, and target-transition counts
are listed in~\cref{tab:appendix-sampled-episodes}.

Transition success is judged by a human evaluator based on the generated command(s), the current state image, and the reference image of the
next state.
A transition is accepted when the evaluator judges that the command(s) would produce the reference next state
sufficiently accurately to continue the procedure.
Folds requiring geometric precision, such as alignment of vertices, edges, center lines,
or reference markers, are judged strictly.
For folds with natural shape variation, a command may be accepted if the evaluator judges that it would produce a state visually
consistent with the intended fold and would not affect subsequent steps.
This protocol measures retry-based correction with the current state image unchanged, not full physical recovery
from a deviated folded state.

\subsection{Task 2 protocol}
\label[appendix]{appendix:detailed-operation-for-task2}

Task 2 uses the same procedures, start--target pairs, LLM controllers, simulator interface, attempt budget, and human judging criteria as Task 1.
However, unlike Task 1, the LLM does not directly receive the reference image sequence, textual procedure, or target state.
Instead, a human instructor has access to the reference procedure and target state, observes the current state image, and provides step-by-step natural language guidance to the LLM.

For each step-specific request, the LLM receives the current simulator image and the corresponding human instruction from the instructor.
The LLM is instructed to output exactly one non-empty code block containing only the simulator command or commands for the given step.
If the instruction is ambiguous or underspecified and the LLM cannot determine the command, it may output a natural-language clarification question instead.
The instructor answers, and the LLM may ask further questions.
The LLM eventually outputs a command for the same step.
The instructor may read, paraphrase, or locally clarify the relevant reference instruction.
The instructor must describe the fold in natural, geometric terms rather than referring to the dot stickers or guide lines used for data collection.
However, the instructor may not directly issue simulator commands, physically manipulate the paper, directly operate the environment, or reveal the full future reference sequence in advance.

A command attempt is evaluated in the same way as in Task 1.
If a human evaluator judges the generated command(s) to be correct for reaching the
reference next state, the transition is counted as successful.
Otherwise, the attempt is counted as failed.
When a failed attempt occurs, the instructor provides free-form corrective feedback
instead of the fixed feedback messages used in Task 1.
If an attempt is judged unsuccessful by a human evaluator, the simulator keeps
the current state image unchanged, and the LLM retries the same transition using the
instructor's corrective feedback.

Each transition allows at most three command attempts.
If the transition cannot be completed within this attempt budget, the episode terminates
and is marked as failed.
An episode is counted as successful only if all target transitions from the sampled start
state to the target state are completed within the attempt budget.
Each sampled episode is evaluated once.

Each episode was handled by one instructor.
In this initial study, the instructor for each episode also served as the evaluator for that episode.
The evaluator judged the generated command(s) using the same criteria as in Task 1.
Although the same person served as instructor and evaluator, an independent third-party evaluator re-judged the commands generated for the transitions of the same 50 sampled origami designs in both tasks and agreed with the original judgments 86.3\% of the time.
Instructor instructions, corrective feedback, LLM-generated commands, current state images and command visualizations, and evaluator judgments were recorded.
As in Task 1, this protocol evaluates retry-based correction with the current state image unchanged and does not measure full recovery from a physically deviated folded state.

\section{Benchmark simulation details}
\label[appendix]{appendix:benchmark-simulation-details}

This appendix provides reproducibility details for the simulator-based benchmark experiments
in~\cref{sec:Benchmark-Simulation}.
All artifacts needed to reproduce or verify the evaluation are available at:
\url{https://github.com/tohoku-nlp/OrigamiTrace}.

\subsection{LLM agent configuration}
\label[appendix]{appendix:llm-agent-configuration}

We used three multimodal LLMs as controllers in both Task 1 and Task 2, each accessed directly through its API.
The same three LLMs were used for all benchmark simulation experiments.
The conversation was reset at the beginning of each episode unless otherwise stated.

\paragraph{Model information.}
We used three multimodal LLM controllers as agents for our evaluation: GPT 5.6 Sol, Gemma 4 31B, and Gemma 3 12B IT.

\paragraph{Access date and interface.}
We evaluated the agents from September 20 to November 5, 2026, accessing each model directly through its API rather than a chat interface: GPT 5.6 Sol through the OpenAI API, and Gemma 4 31B and Gemma 3 12B IT through Amazon Bedrock.

\paragraph{Decoding settings.}
For GPT 5.6 Sol, we set the reasoning effort to low and allowed up to 32{,}000 completion tokens.
For Gemma 3 12B IT, we set the temperature to 0.
For Gemma 4 31B, the API only accepts its default temperature.
For both Gemma models, the maximum number of output tokens was set to 2{,}048.
All other decoding settings were left at each API's default.

\subsection{Episode sampling}
\label[appendix]{appendix:episode-sampling}

We used all 348 reference origami procedures in the dataset.
For each procedure, we randomly selected a start state and set the target state at most 10 transitions later, or at the final state if fewer than 10 transitions remained.
The same 348 procedures and the same start--target pairs were used for Task 1 and Task 2.
However, the actually attempted transitions may differ between the two tasks because an episode terminates once any transition cannot be completed within the attempt budget.
We limited the maximum number of target transitions to 10 because evaluating full procedures would make the episode success rate too sparse for an initial study.

\paragraph{Sampled episode list.}
\begin{center}
\scriptsize
\tabcolsep 2pt
\begin{longtable}{rp{0.37\linewidth}rrrr}
\caption{
Sampled episodes used in the benchmark simulation.
The ID corresponds to the numeric prefix of the folder and file names in the released dataset (e.g., \texttt{001\_Long-Horned\_Samurai\_Helmet}).
}
\label{tab:appendix-sampled-episodes} \\
\toprule
ID & Design name & Start & Target & \# Trans. & Segment bin \\
\midrule
\endfirsthead
\multicolumn{6}{l}{\tablename\ \thetable{} -- continued from previous page} \\
\toprule
ID & Design name & Start & Target & \# Trans. & Segment bin \\
\midrule
\endhead
\midrule
\multicolumn{6}{r}{Continued on next page} \\
\endfoot
\bottomrule
\endlastfoot
1 & Long-Horned Samurai Helmet & 5 & 13 & 8 & 7--10 \\
2 & Boat & 6 & 16 & 10 & 7--10 \\
3 & Yacht & 7 & 8 & 1 & 1--3 \\
4 & Hopping Frog & 20 & 25 & 5 & 4--6 \\
5 & Camera & 6 & 16 & 10 & 7--10 \\
6 & Chomper & 5 & 15 & 10 & 7--10 \\
7 & Iris & 51 & 53 & 2 & 1--3 \\
8 & Medal & 12 & 22 & 10 & 7--10 \\
9 & Butterfly & 21 & 31 & 10 & 7--10 \\
10 & Sanbo (Offering Stand) & 5 & 15 & 10 & 7--10 \\
11 & Frog & 50 & 57 & 7 & 7--10 \\
12 & Treasure Ship & 13 & 23 & 10 & 7--10 \\
13 & Box & 14 & 17 & 3 & 1--3 \\
14 & Balloon & 14 & 22 & 8 & 7--10 \\
15 & Samurai's Attendant & 9 & 14 & 5 & 4--6 \\
16 & Brush Washer & 6 & 16 & 10 & 7--10 \\
17 & Square-Folded Letter & 15 & 22 & 7 & 7--10 \\
18 & Paper Popper & 6 & 9 & 3 & 1--3 \\
19 & Sumo Wrestler & 12 & 16 & 4 & 4--6 \\
20 & Traditional Japanese pleated trousers & 4 & 14 & 10 & 7--10 \\
21 & Pig & 22 & 24 & 2 & 1--3 \\
22 & Penguin & 7 & 17 & 10 & 7--10 \\
23 & House & 3 & 9 & 6 & 4--6 \\
24 & Dog & 3 & 5 & 2 & 1--3 \\
25 & Lipstick Case & 12 & 15 & 3 & 1--3 \\
26 & Fox & 10 & 20 & 10 & 7--10 \\
27 & Hungry Crow & 3 & 13 & 10 & 7--10 \\
28 & Double-Crested Samurai Helmet & 1 & 11 & 10 & 7--10 \\
29 & Chinese phoenix & 26 & 32 & 6 & 4--6 \\
30 & Ring & 15 & 25 & 10 & 7--10 \\
31 & Airplane & 10 & 15 & 5 & 4--6 \\
32 & Airplane & 6 & 13 & 7 & 7--10 \\
33 & Bellflower & 8 & 18 & 10 & 7--10 \\
34 & Kite Airplane & 2 & 12 & 10 & 7--10 \\
35 & Navel Paper Plane & 8 & 15 & 7 & 7--10 \\
36 & Persimmon & 26 & 29 & 3 & 1--3 \\
37 & Snail & 32 & 42 & 10 & 7--10 \\
38 & Rose & 6 & 16 & 10 & 7--10 \\
39 & Motorboat & 23 & 25 & 2 & 1--3 \\
40 & Jar & 3 & 13 & 10 & 7--10 \\
41 & Catfish & 14 & 19 & 5 & 4--6 \\
42 & Plump Sparrow & 21 & 22 & 1 & 1--3 \\
43 & Many Faces & 9 & 16 & 7 & 7--10 \\
44 & Cat & 6 & 8 & 2 & 1--3 \\
45 & Crane Gift Envelope & 16 & 26 & 10 & 7--10 \\
46 & Pigeon & 6 & 11 & 5 & 4--6 \\
47 & Celebration Turtle & 21 & 31 & 10 & 7--10 \\
48 & Celebration Crane & 12 & 22 & 10 & 7--10 \\
49 & Wristwatch & 4 & 14 & 10 & 7--10 \\
50 & Dahlia & 8 & 18 & 10 & 7--10 \\
51 & Spear plane & 5 & 12 & 7 & 7--10 \\
52 & Table & 16 & 26 & 10 & 7--10 \\
53 & Table & 9 & 19 & 10 & 7--10 \\
54 & Chair & 8 & 18 & 10 & 7--10 \\
55 & Sea Lion/Fur Seal & 10 & 20 & 10 & 7--10 \\
56 & Red-and-White Crane & 5 & 15 & 10 & 7--10 \\
57 & Cat & 8 & 10 & 2 & 1--3 \\
58 & Dog & 7 & 10 & 3 & 1--3 \\
59 & Boat & 11 & 12 & 1 & 1--3 \\
60 & Cicada & 9 & 12 & 3 & 1--3 \\
61 & Traditional Japanese Festival Coat & 2 & 12 & 10 & 7--10 \\
62 & Lotus Flower & 28 & 34 & 6 & 4--6 \\
63 & Fish & 4 & 6 & 2 & 1--3 \\
64 & Hat & 3 & 13 & 10 & 7--10 \\
65 & Crane & 4 & 14 & 10 & 7--10 \\
66 & Boots & 4 & 14 & 10 & 7--10 \\
67 & Wallet & 5 & 15 & 10 & 7--10 \\
68 & Wallet & 4 & 7 & 3 & 1--3 \\
69 & 3D Snail & 28 & 38 & 10 & 7--10 \\
70 & Ring & 6 & 16 & 10 & 7--10 \\
71 & Shirt & 5 & 15 & 10 & 7--10 \\
72 & Airplane & 5 & 14 & 9 & 7--10 \\
73 & Paper Airplane & 1 & 11 & 10 & 7--10 \\
74 & Jet Plane & 4 & 12 & 8 & 7--10 \\
75 & Dried-Squid Airplane & 6 & 12 & 6 & 4--6 \\
76 & Paper Airplane & 3 & 13 & 10 & 7--10 \\
77 & Rose & 11 & 21 & 10 & 7--10 \\
78 & Squid Airplane & 3 & 11 & 8 & 7--10 \\
79 & Carp Streamer & 1 & 11 & 10 & 7--10 \\
80 & Tall Samurai Helmet & 11 & 12 & 1 & 1--3 \\
81 & Samurai Helmet & 6 & 10 & 4 & 4--6 \\
82 & Leaf/Young Leaf & 18 & 20 & 2 & 1--3 \\
83 & Little Bird & 7 & 16 & 9 & 7--10 \\
84 & Snow Rabbit & 2 & 12 & 10 & 7--10 \\
85 & Balloon Rabbit & 10 & 20 & 10 & 7--10 \\
86 & Rabbit & 4 & 14 & 10 & 7--10 \\
87 & Rabbit Face & 2 & 8 & 6 & 4--6 \\
88 & Ring & 14 & 24 & 10 & 7--10 \\
89 & Pinwheel & 4 & 14 & 10 & 7--10 \\
90 & Spinning Pinwheel & 1 & 11 & 10 & 7--10 \\
91 & Hina Doll (Girl's Day Dolls) & 6 & 12 & 6 & 4--6 \\
92 & Hina Doll (Girl's Day Dolls) & 2 & 12 & 10 & 7--10 \\
93 & Hina Doll (Girl's Day Dolls) & 7 & 9 & 2 & 1--3 \\
94 & Sanbo (Offering Stand) with Legs & 16 & 26 & 10 & 7--10 \\
95 & Candy Bowl & 5 & 15 & 10 & 7--10 \\
96 & Horned Box & 16 & 26 & 10 & 7--10 \\
97 & Horned Box & 15 & 25 & 10 & 7--10 \\
98 & Simple Box & 11 & 21 & 10 & 7--10 \\
99 & Simple Box & 31 & 32 & 1 & 1--3 \\
100 & Simple Box & 15 & 16 & 1 & 1--3 \\
101 & Masu (Wooden Sake Cup) & 12 & 22 & 10 & 7--10 \\
102 & Tatou (Paper folder with flaps) Thread Case & 19 & 22 & 3 & 1--3 \\
103 & Tatou (Paper folder with flaps) Case & 8 & 13 & 5 & 4--6 \\
104 & Simple Box & 15 & 20 & 5 & 4--6 \\
105 & Hexagonal Tatou (Paper folder with flaps) & 18 & 26 & 8 & 7--10 \\
106 & Octagonal Tatou (Paper folder with flaps) & 10 & 20 & 10 & 7--10 \\
107 & Tatou (Paper folder with flaps) Case & 20 & 30 & 10 & 7--10 \\
108 & Hagoita (Wooden) Paddle & 3 & 10 & 7 & 7--10 \\
109 & Camellia & 6 & 16 & 10 & 7--10 \\
110 & Bellflower & 4 & 14 & 10 & 7--10 \\
111 & Crane Gift Envelope & 4 & 14 & 10 & 7--10 \\
112 & Origami Crane & 5 & 15 & 10 & 7--10 \\
113 & Cup & 5 & 8 & 3 & 1--3 \\
114 & Lion-Dance Finger Puppet & 20 & 23 & 3 & 1--3 \\
115 & Animal Finger Puppet & 5 & 13 & 8 & 7--10 \\
116 & Spinning Butterfly & 5 & 10 & 5 & 4--6 \\
117 & Candy Box & 11 & 21 & 10 & 7--10 \\
118 & Carrier Pigeon & 1 & 11 & 10 & 7--10 \\
119 & Moving-Mouth Kappa (Japanese River Imp) Puppet & 8 & 9 & 1 & 1--3 \\
120 & Ribbon & 11 & 14 & 3 & 1--3 \\
121 & Parrot & 8 & 18 & 10 & 7--10 \\
122 & Dolphin & 3 & 13 & 10 & 7--10 \\
123 & Grasshopper & 9 & 19 & 10 & 7--10 \\
124 & Shrimp & 7 & 17 & 10 & 7--10 \\
125 & Chicken & 7 & 12 & 5 & 4--6 \\
126 & Seismosaurus & 26 & 32 & 6 & 4--6 \\
127 & Plesiosaurus & 27 & 37 & 10 & 7--10 \\
128 & Brachiosaurus & 18 & 28 & 10 & 7--10 \\
129 & Wild Boar & 2 & 12 & 10 & 7--10 \\
130 & Waterfowl & 2 & 10 & 8 & 7--10 \\
131 & Mammoth & 44 & 54 & 10 & 7--10 \\
132 & Spinosaurus & 15 & 25 & 10 & 7--10 \\
133 & Pteranodon & 5 & 15 & 10 & 7--10 \\
134 & Triceratops & 15 & 25 & 10 & 7--10 \\
135 & Tyrannosaurus & 12 & 22 & 10 & 7--10 \\
136 & Dragon & 31 & 41 & 10 & 7--10 \\
137 & Monkey & 16 & 22 & 6 & 4--6 \\
138 & Elephant & 4 & 14 & 10 & 7--10 \\
139 & Duck & 7 & 17 & 10 & 7--10 \\
140 & Peacock & 17 & 27 & 10 & 7--10 \\
141 & Swan & 16 & 20 & 4 & 4--6 \\
142 & Crow & 10 & 20 & 10 & 7--10 \\
143 & Parakeet & 13 & 16 & 3 & 1--3 \\
144 & Panda & 8 & 18 & 10 & 7--10 \\
145 & Squirrel & 16 & 26 & 10 & 7--10 \\
146 & Penguin & 10 & 20 & 10 & 7--10 \\
147 & Mount Fuji & 20 & 30 & 10 & 7--10 \\
148 & Four-Leaf Clover & 10 & 20 & 10 & 7--10 \\
149 & Acorn & 3 & 13 & 10 & 7--10 \\
150 & Japanese Maple & 34 & 40 & 6 & 4--6 \\
151 & Ribbon & 4 & 14 & 10 & 7--10 \\
152 & Lantern & 7 & 17 & 10 & 7--10 \\
153 & Handbag & 6 & 13 & 7 & 7--10 \\
154 & Leek & 6 & 16 & 10 & 7--10 \\
155 & Daikon Radish & 2 & 12 & 10 & 7--10 \\
156 & Mushroom & 16 & 20 & 4 & 4--6 \\
157 & Chestnut & 6 & 16 & 10 & 7--10 \\
158 & Shaved Ice & 24 & 25 & 1 & 1--3 \\
159 & Kagami Mochi (Mirror Rice Cake) & 11 & 21 & 10 & 7--10 \\
160 & Dress & 12 & 22 & 10 & 7--10 \\
161 & One-Piece Dress & 27 & 29 & 2 & 1--3 \\
162 & Necktie & 4 & 14 & 10 & 7--10 \\
163 & Shirt & 9 & 17 & 8 & 7--10 \\
164 & Jacket & 4 & 14 & 10 & 7--10 \\
165 & Kimono (Traditional Japanese Garment) & 3 & 13 & 10 & 7--10 \\
166 & Witch Hat & 7 & 17 & 10 & 7--10 \\
167 & Reindeer & 22 & 26 & 4 & 4--6 \\
168 & Koma (Spinning Top) & 4 & 8 & 4 & 4--6 \\
169 & Koma (Spinning Top) & 6 & 16 & 10 & 7--10 \\
170 & Ema (Wooden prayer plaque) & 1 & 11 & 10 & 7--10 \\
171 & Musical Note & 12 & 13 & 1 & 1--3 \\
172 & Heart & 6 & 10 & 4 & 4--6 \\
173 & Star & 13 & 21 & 8 & 7--10 \\
174 & Sled & 19 & 22 & 3 & 1--3 \\
175 & Candle & 23 & 28 & 5 & 4--6 \\
176 & Candle & 12 & 16 & 4 & 4--6 \\
177 & Candle & 16 & 17 & 1 & 1--3 \\
178 & Gift Box & 1 & 11 & 10 & 7--10 \\
179 & Bell & 4 & 13 & 9 & 7--10 \\
180 & Bell & 10 & 13 & 3 & 1--3 \\
181 & Bell & 1 & 11 & 10 & 7--10 \\
182 & Space Shuttle & 2 & 12 & 10 & 7--10 \\
183 & Glider & 11 & 16 & 5 & 4--6 \\
184 & Daruma (Goal-setting Doll) & 15 & 20 & 5 & 4--6 \\
185 & Snake & 5 & 15 & 10 & 7--10 \\
186 & Pumpkin & 3 & 13 & 10 & 7--10 \\
187 & Ghost & 1 & 11 & 10 & 7--10 \\
188 & Rabbit & 28 & 38 & 10 & 7--10 \\
189 & Flapping Bird & 2 & 12 & 10 & 7--10 \\
190 & Basket & 39 & 41 & 2 & 1--3 \\
191 & Snow Child & 13 & 19 & 6 & 4--6 \\
192 & Sycee & 3 & 13 & 10 & 7--10 \\
193 & Dollar Boots & 5 & 15 & 10 & 7--10 \\
194 & Cute Letter & 4 & 9 & 5 & 4--6 \\
195 & Cute Letter & 11 & 16 & 5 & 4--6 \\
196 & Cute Letter & 8 & 12 & 4 & 4--6 \\
197 & Heart Letter & 2 & 12 & 10 & 7--10 \\
198 & Ladybug & 8 & 11 & 3 & 1--3 \\
199 & Chopstick Sleeve & 9 & 12 & 3 & 1--3 \\
200 & Teacup & 4 & 10 & 6 & 4--6 \\
201 & Jumping Snake & 6 & 16 & 10 & 7--10 \\
202 & Truck & 3 & 10 & 7 & 7--10 \\
203 & Rocket & 5 & 10 & 5 & 4--6 \\
204 & Convertible & 5 & 12 & 7 & 7--10 \\
205 & Car & 5 & 10 & 5 & 4--6 \\
206 & Car & 8 & 11 & 3 & 1--3 \\
207 & Baby Dinosaur & 6 & 16 & 10 & 7--10 \\
208 & Parent-and-Child Turtle & 6 & 11 & 5 & 4--6 \\
209 & Parent-and-Child Turtle & 10 & 20 & 10 & 7--10 \\
210 & Parent-and-Child Penguins & 8 & 15 & 7 & 7--10 \\
211 & Four Frogs & 21 & 31 & 10 & 7--10 \\
212 & Piggyback Bird & 4 & 14 & 10 & 7--10 \\
213 & Parent-and-Child Swans & 4 & 14 & 10 & 7--10 \\
214 & Hatching Chick & 16 & 21 & 5 & 4--6 \\
215 & Seaweed & 7 & 17 & 10 & 7--10 \\
216 & Seaweed & 11 & 21 & 10 & 7--10 \\
217 & Ginkgo Leaf & 13 & 14 & 1 & 1--3 \\
218 & Morning Glory & 14 & 24 & 10 & 7--10 \\
219 & Lion & 17 & 19 & 2 & 1--3 \\
220 & Lioness & 5 & 13 & 8 & 7--10 \\
221 & Rhinoceros & 6 & 16 & 10 & 7--10 \\
222 & Leopard & 8 & 18 & 10 & 7--10 \\
223 & Alpaca & 3 & 13 & 10 & 7--10 \\
224 & Water Buffalo & 20 & 30 & 10 & 7--10 \\
225 & Kangaroo & 24 & 27 & 3 & 1--3 \\
226 & Ogre Face & 44 & 50 & 6 & 4--6 \\
227 & Lucky Crane & 32 & 42 & 10 & 7--10 \\
228 & Camel & 14 & 22 & 8 & 7--10 \\
229 & Playful Cat & 12 & 22 & 10 & 7--10 \\
230 & White Egret & 11 & 21 & 10 & 7--10 \\
231 & Owl & 1 & 11 & 10 & 7--10 \\
232 & Shoebill & 4 & 14 & 10 & 7--10 \\
233 & Duck & 4 & 14 & 10 & 7--10 \\
234 & Flying Duck & 13 & 17 & 4 & 4--6 \\
235 & Japanese Bush Warbler & 33 & 37 & 4 & 4--6 \\
236 & Peacock & 5 & 15 & 10 & 7--10 \\
237 & Dragonfly & 20 & 27 & 7 & 7--10 \\
238 & Ramune (Japanese Marble) Soda & 4 & 14 & 10 & 7--10 \\
239 & Shaved Ice & 14 & 23 & 9 & 7--10 \\
240 & Shaved Ice & 9 & 13 & 4 & 4--6 \\
241 & Taiyaki (Fish-shaped Cake) & 3 & 11 & 8 & 7--10 \\
242 & Soft-Serve Ice Cream & 11 & 14 & 3 & 1--3 \\
243 & Drinking Straw & 14 & 15 & 1 & 1--3 \\
244 & Shortcake (Sponge Cake with Whipped Cream) & 5 & 14 & 9 & 7--10 \\
245 & Croissant & 2 & 12 & 10 & 7--10 \\
246 & Plate & 15 & 21 & 6 & 4--6 \\
247 & Kitchen Knife & 8 & 14 & 6 & 4--6 \\
248 & Sailor Suit & 3 & 7 & 4 & 4--6 \\
249 & Skirt & 5 & 11 & 6 & 4--6 \\
250 & Skirt & 8 & 18 & 10 & 7--10 \\
251 & T-Shirt & 1 & 8 & 7 & 7--10 \\
252 & One-Piece Dress & 11 & 13 & 2 & 1--3 \\
253 & One-Piece Dress & 9 & 13 & 4 & 4--6 \\
254 & One-Piece Dress & 3 & 11 & 8 & 7--10 \\
255 & One-Piece Dress & 9 & 12 & 3 & 1--3 \\
256 & One-Piece Dress & 4 & 14 & 10 & 7--10 \\
257 & One-Piece Dress & 10 & 15 & 5 & 4--6 \\
258 & One-Piece Dress & 11 & 18 & 7 & 7--10 \\
259 & Raincoat & 5 & 14 & 9 & 7--10 \\
260 & Polo Shirt & 11 & 14 & 3 & 1--3 \\
261 & Swimsuit & 2 & 12 & 10 & 7--10 \\
262 & Coat & 9 & 11 & 2 & 1--3 \\
263 & Necktie & 8 & 12 & 4 & 4--6 \\
264 & Double Heart & 18 & 20 & 2 & 1--3 \\
265 & Broken Heart & 13 & 23 & 10 & 7--10 \\
266 & Heart Crane & 20 & 30 & 10 & 7--10 \\
267 & Winged Heart & 6 & 15 & 9 & 7--10 \\
268 & Two-Tone Heart & 5 & 15 & 10 & 7--10 \\
269 & Two-Tone Heart & 11 & 18 & 7 & 7--10 \\
270 & Kappa (Japanese Water Imp) & 10 & 20 & 10 & 7--10 \\
271 & Skeleton & 3 & 12 & 9 & 7--10 \\
272 & Dracula Face & 6 & 16 & 10 & 7--10 \\
273 & Witch Face & 8 & 11 & 3 & 1--3 \\
274 & Bat & 11 & 15 & 4 & 4--6 \\
275 & Bat & 3 & 9 & 6 & 4--6 \\
276 & Number 1 & 3 & 13 & 10 & 7--10 \\
277 & Number 2 & 10 & 20 & 10 & 7--10 \\
278 & Number 3 & 2 & 12 & 10 & 7--10 \\
279 & Number 4 & 5 & 12 & 7 & 7--10 \\
280 & Number 5 & 10 & 18 & 8 & 7--10 \\
281 & Number 6 & 15 & 17 & 2 & 1--3 \\
282 & Number 7 & 3 & 13 & 10 & 7--10 \\
283 & Number 8 & 15 & 18 & 3 & 1--3 \\
284 & Number 9 & 1 & 11 & 10 & 7--10 \\
285 & Trophy & 19 & 22 & 3 & 1--3 \\
286 & Parasaurolophus & 5 & 15 & 10 & 7--10 \\
287 & Buddhist Monk & 6 & 16 & 10 & 7--10 \\
288 & Dwarf & 11 & 18 & 7 & 7--10 \\
289 & Tengu (long-nosed goblin) & 7 & 17 & 10 & 7--10 \\
290 & 3D Rabbit & 17 & 27 & 10 & 7--10 \\
291 & Simple Star & 22 & 23 & 1 & 1--3 \\
292 & Cute Bird & 12 & 22 & 10 & 7--10 \\
293 & Sword & 21 & 27 & 6 & 4--6 \\
294 & Hummingbird & 10 & 16 & 6 & 4--6 \\
295 & Fish & 2 & 12 & 10 & 7--10 \\
296 & Glasses & 11 & 13 & 2 & 1--3 \\
297 & Book & 20 & 21 & 1 & 1--3 \\
298 & Twin Boat & 16 & 26 & 10 & 7--10 \\
299 & Tulip & 7 & 9 & 2 & 1--3 \\
300 & Bear & 8 & 18 & 10 & 7--10 \\
301 & Ogre & 8 & 13 & 5 & 4--6 \\
302 & Single Sneaker & 20 & 30 & 10 & 7--10 \\
303 & Ladybug & 4 & 11 & 7 & 7--10 \\
304 & Organ & 2 & 12 & 10 & 7--10 \\
305 & Alphabet A & 4 & 8 & 4 & 4--6 \\
306 & Alphabet B & 10 & 11 & 1 & 1--3 \\
307 & Alphabet C & 1 & 11 & 10 & 7--10 \\
308 & Alphabet D & 4 & 8 & 4 & 4--6 \\
309 & Alphabet E & 7 & 8 & 1 & 1--3 \\
310 & Alphabet F & 6 & 10 & 4 & 4--6 \\
311 & Alphabet G & 17 & 18 & 1 & 1--3 \\
312 & Alphabet H & 5 & 7 & 2 & 1--3 \\
313 & Alphabet I & 1 & 6 & 5 & 4--6 \\
314 & Alphabet J & 5 & 10 & 5 & 4--6 \\
315 & Alphabet K & 5 & 7 & 2 & 1--3 \\
316 & Alphabet L & 1 & 11 & 10 & 7--10 \\
317 & Alphabet M & 3 & 6 & 3 & 1--3 \\
318 & Alphabet N & 8 & 11 & 3 & 1--3 \\
319 & Alphabet O & 7 & 10 & 3 & 1--3 \\
320 & Alphabet P & 2 & 10 & 8 & 7--10 \\
321 & Alphabet Q & 12 & 18 & 6 & 4--6 \\
322 & Alphabet R & 7 & 9 & 2 & 1--3 \\
323 & Alphabet S & 8 & 14 & 6 & 4--6 \\
324 & Alphabet T & 6 & 12 & 6 & 4--6 \\
325 & Alphabet U & 7 & 16 & 9 & 7--10 \\
326 & Alphabet V & 5 & 11 & 6 & 4--6 \\
327 & Alphabet W & 9 & 14 & 5 & 4--6 \\
328 & Alphabet X & 1 & 11 & 10 & 7--10 \\
329 & Alphabet Y & 4 & 14 & 10 & 7--10 \\
330 & Alphabet Z & 13 & 15 & 2 & 1--3 \\
331 & Minus Sign & 6 & 11 & 5 & 4--6 \\
332 & Plus Sign & 9 & 11 & 2 & 1--3 \\
333 & Division Sign & 7 & 16 & 9 & 7--10 \\
334 & Equals Sign & 6 & 10 & 4 & 4--6 \\
335 & Samurai & 3 & 13 & 10 & 7--10 \\
336 & Pheasant & 16 & 18 & 2 & 1--3 \\
337 & Parakeet & 7 & 17 & 10 & 7--10 \\
338 & Eagle & 1 & 11 & 10 & 7--10 \\
339 & Horned Owl & 14 & 24 & 10 & 7--10 \\
340 & Horned Owl & 3 & 13 & 10 & 7--10 \\
341 & Flying Squirrel & 3 & 13 & 10 & 7--10 \\
342 & Little Fox & 5 & 15 & 10 & 7--10 \\
343 & Chimpanzee & 17 & 21 & 4 & 4--6 \\
344 & Anglerfish & 17 & 23 & 6 & 4--6 \\
345 & Turban Shell & 2 & 12 & 10 & 7--10 \\
346 & Softshell Turtle & 17 & 22 & 5 & 4--6 \\
347 & Hermit Crab & 8 & 18 & 10 & 7--10 \\
348 & Whale & 7 & 13 & 6 & 4--6 \\
\midrule
\multicolumn{4}{r}{Total} & 2505 & \\
\end{longtable}
\end{center}

\subsection{Episode initialization}
\label[appendix]{appendix:episode-initialization}

The simulator represents each paper state by the corresponding dataset image: an episode starts from the image of the sampled start state, and a command judged correct by a human evaluator advances the state to the image of the next reference state.
If an attempt is judged unsuccessful by the evaluator, the simulator keeps the current state image unchanged, and the agent is prompted to retry the same transition.
The paper is a single sheet, not necessarily square, and the dot stickers and radial guide lines remain visible; physical contact conditions are not modeled.
The agent receives no explicit description of layer ordering, existing creases, or orientation, and must infer them from the image.

\subsection{Simulator interface}
\label[appendix]{appendix:simulator-interface}

Simulator commands were represented in JSON format.
The command interface used in the experiments included \texttt{fold},
\texttt{make\_crease}, and \texttt{rotate\_3d}.
A \texttt{fold} command may optionally also tuck the folded flap into a pocket or
layer gap, by specifying a target point and the pair of visible layers surrounding it,
instead of issuing insertion as a separate command.
The interface also includes a \texttt{finish} command for future extensions, but
\texttt{finish} was not used in the present experiments because trial termination was
determined by evaluator judgment.
An example visualization of a command, showing its source point, target point, and crease
line, is given in~\cref{fig:task1-judging-interface}.

\paragraph{Command specification.}
The simulator command interface was specified in \texttt{command\_specification.json}.
The full command specification is included in the reproduction artifacts at
\url{https://github.com/tohoku-nlp/OrigamiTrace}.
The file defines the coordinate system, command invocation and return formats, fold
specification schema, simulator conventions, and supported commands.
An excerpt is shown below.

\begingroup
\small
\begin{verbatim}
{
  "global_settings": {
    "coordinate_system": "normalized_current_view",
    "origin": "current_projected_center_of_mass",
    "unit": "current_view_max_xy_extent_from_center",
    "angle_unit": "degree",
    "layer_order": "visible_top_to_bottom",
    "view_axes": {
      "x_axis": "rightward in the current view",
      "y_axis": "upward in the current view",
      "z_axis": "toward the viewer"
    }
  },

  "command_invocation_format": {
    "description": "Each command is represented by command_name and parameters.",
    "format": {
      "command_name": "String",
      "parameters": "Object"
    }
  },

  "fold_info_schema": {
    "optional_fields": [
      {"name": "pocket_point", "type": "List[Number, Number]"},
      {"name": "destination_layer", "type": "List[Integer, Integer]"}
    ]
  },

  "commands": [
    {"command_name": "fold"},
    {"command_name": "make_crease"},
    {"command_name": "rotate_3d"},
    {"command_name": "finish"}
  ]
}
\end{verbatim}
\endgroup

\paragraph{Representative command examples.}
We also prepared a JSON file containing representative simulator-command examples,
covering common operations such as full folds, partial folds, precreases, simultaneous
folds, insertions, and 3D rotations.
The full example file is included in the reproduction artifacts at
\url{https://github.com/tohoku-nlp/OrigamiTrace}.
An excerpt is shown below.

\begingroup
\small
\begin{verbatim}
{
  "examples": [
    {
      "name": "vertical_full_valley_fold_all_layers",
      "covers": [
        "valley fold",
        "half fold",
        "flat fold",
        "all-layer fold"
      ],
      "commands": [
        {
          "command_name": "fold",
          "parameters": {
            "fold_infos": [
              {
                "crease_line_edge1": [0, -0.9],
                "crease_line_edge2": [0, 0.9],
                "source_point": [0.8, 0],
                "target_point": [-0.8, 0],
                "source_layers": ["all"],
                "theta": 180,
                "fold_type": "valley"
              }
            ]
          }
        }
      ]
    },

    {
      "name": "horizontal_full_mountain_fold_top_layer",
      "covers": [
        "mountain fold",
        "top-layer fold"
      ],
      "commands": [
        {
          "command_name": "fold",
          "parameters": {
            "fold_infos": [
              {
                "crease_line_edge1": [-0.9, 0],
                "crease_line_edge2": [0.9, 0],
                "source_point": [0, 0.7],
                "target_point": [0, -0.7],
                "source_layers": [1],
                "theta": 180,
                "fold_type": "mountain"
              }
            ]
          }
        }
      ]
    }
  ]
}
\end{verbatim}
\endgroup

\subsection{Prompts and feedback}
\label[appendix]{appendix:prompts-and-feedback}

The LLM was prompted to generate structured simulator commands in JSON format.
The exact prompt differed between Task 1 and Task 2 because the information available
to the agent differed.

\paragraph{Task 1 prompt.}
In Task 1, the LLM received the reference image sequence and the CSV file containing the
textual entries.
For each step-specific request, the model was given the current reference image and the
corresponding textual instruction, and was required to output only simulator commands.
The system prompt used for Task 1 is shown below.

\begingroup
\small
\begin{Verbatim}[breaklines=true, breakanywhere=true, breaksymbolleft={}, breaksymbolright={}]
# Instruction
You are an execution agent that operates an origami simulator. Your task is to convert the specified origami step into simulator commands according to the available command specifications. The user will first provide all reference images as separate image files and one CSV file containing all textual instructions. After that, for each step, the user will provide the current reference image for that step and the corresponding textual instruction from the CSV. Generate the required simulator command or commands based on the provided current reference image and textual instruction.

# Input format
The initial input consists of all reference images as separate image files and one CSV file containing all textual instructions.

Each following input consists of:
1. The current reference image for the step to execute
2. The corresponding textual instruction from the CSV

# Output format
For step-specific requests, output exactly one non-empty code block. Do not include any explanation or text outside the code block. The code block must contain only the simulator command or commands for the specified step. The code block must contain strict JSON.
For the initial input files, output exactly `OK`.
\end{Verbatim}
\endgroup

\paragraph{Task 2 prompt.}
In Task 2, the LLM did not receive the reference image sequence, textual procedure, or
target state directly.
For each step-specific request, the model was given the current simulator image and a
human instruction from the instructor, and was required to output either the simulator
commands for that step or, if the instruction was ambiguous or underspecified, a single
natural-language clarification question.
The system prompt used for Task 2 is shown below.

\begingroup
\small
\begin{Verbatim}[breaklines=true, breakanywhere=true, breaksymbolleft={}, breaksymbolright={}]
# Instruction
You are an execution agent that operates an origami simulator. Your task is to convert the given origami step into simulator commands according to the available command specifications.
For each request, the user will provide the current origami state as an image and a human instruction for that state.
Use these to generate the required simulator command or commands.

# Input format
Each input consists of one image showing the current origami state and one corresponding human instruction.

# Output format
Respond in exactly one of the following two ways:
- If you can determine the command: output exactly one non-empty code block containing only the simulator command or commands for the given step. Do not include any explanation or text outside the code block. The code block must contain strict JSON.
- If the instruction is ambiguous or underspecified and you cannot determine the command: do not output a code block. Instead, output exactly one clarification question in natural language. Phrase the question using only natural geometric description — edges, corners, midpoints, existing creases, directions, angles, and fractions of a length. Do not refer to the colored sticker markers on the paper, or to the guide lines on the surface beneath the paper.
Never output both a code block and a question in the same response.
\end{Verbatim}
\endgroup

\paragraph{Failure feedback.}
When a command attempt failed in Task 1, the LLM received one of the following fixed
feedback messages:
\begin{verbatim}
The command is invalid. Please regenerate it.
The fold is incorrect. Please regenerate it.
\end{verbatim}
When a command attempt failed in Task 2, the human instructor provided free-form
corrective feedback.

\subsection{Metric definitions}
\label[appendix]{appendix:metric-definitions}

This section defines the metrics reported in~\cref{tab:benchmark_simulation_results}.
The denominators differ across metrics because an episode terminates once any transition
cannot be completed within the attempt budget.

Let $N_{\mathrm{ep}}$ be the number of sampled episodes, and let $N_{\mathrm{succ\_ep}}$
be the number of episodes in which the target state was reached.
Episode success rate is defined as
\[
\mathrm{Episode\text{-}SR}
=
\frac{N_{\mathrm{succ\_ep}}}{N_{\mathrm{ep}}}.
\]

Let $N_{\mathrm{target\_trans}}$ be the number of target transitions in all sampled
segments, including later transitions in episodes that terminated early.
Let $N_{\mathrm{succ\_trans}}$ be the number of target transitions successfully completed
before termination.
Transition success rate is defined as
\[
\mathrm{Trans\text{-}SR}
=
\frac{N_{\mathrm{succ\_trans}}}{N_{\mathrm{target\_trans}}}.
\]

Let $N_{\mathrm{attempted\_trans}}$ be the number of target transitions actually attempted
before episode termination.
Let $N_{\mathrm{first\_succ}}$ be the number of attempted transitions solved on the first
command attempt.
First-attempt success rate is defined as
\[
\mathrm{First\text{-}Att\text{-}SR}
=
\frac{N_{\mathrm{first\_succ}}}{N_{\mathrm{attempted\_trans}}}.
\]

Let $N_{\mathrm{cmd\_attempts}}$ be the total number of command attempts issued for
attempted target transitions.
Average attempts per transition is defined as
\[
\mathrm{Avg\ Att/Trans}
=
\frac{N_{\mathrm{cmd\_attempts}}}{N_{\mathrm{attempted\_trans}}}.
\]

Let $N_{\mathrm{failed\_first}}$ be the number of transitions whose first command attempt
failed.
Let $N_{\mathrm{retry\_succ}}$ be the number of these transitions that were successfully
completed within the remaining retry budget.
Retry-correction success rate is defined as
\[
\mathrm{Retry\text{-}SR}
=
\frac{N_{\mathrm{retry\_succ}}}{N_{\mathrm{failed\_first}}}.
\]

\paragraph{Metric values in the main experiments.}
\cref{tab:appendix-metric-counts} summarizes the raw counts used to compute the metrics
reported in~\cref{tab:benchmark_simulation_results}.

\begin{table}[t]
\centering
\small
\setlength{\tabcolsep}{3pt}
\caption{
Raw counts used for the metric calculations in the main benchmark simulation experiments.
Sol, G4, and G3 denote GPT 5.6 Sol, Gemma 4 31B, and Gemma 3 12B IT, respectively.
}
\label{tab:appendix-metric-counts}
\begin{tabular}{ll rrr rrr}
\toprule
& & \multicolumn{3}{c}{Task 1} & \multicolumn{3}{c}{Task 2} \\
\cmidrule(lr){3-5}
\cmidrule(lr){6-8}
Symbol & Description & Sol & G4 & G3 & Sol & G4 & G3 \\
\midrule
$N_{\mathrm{ep}}$ & Sampled episodes & 348 & 348 & 348 & 348 & 348 & 348 \\
$N_{\mathrm{succ\_ep}}$ & Successful episodes & 79 & 13 & 6 & 89 & 12 & 10 \\
$N_{\mathrm{target\_trans}}$ & Target transitions & 2{,}505 & 2{,}505 & 2{,}505 & 2{,}505 & 2{,}505 & 2{,}505 \\
$N_{\mathrm{succ\_trans}}$ & Successful target transitions & 855 & 160 & 19 & 1{,}010 & 212 & 76 \\
$N_{\mathrm{attempted\_trans}}$ & Attempted target transitions & 1{,}124 & 495 & 361 & 1{,}269 & 548 & 414 \\
$N_{\mathrm{first\_succ}}$ & First-attempt successes & 689 & 130 & 10 & 742 & 157 & 59 \\
$N_{\mathrm{cmd\_attempts}}$ & Command attempts & 1{,}888 & 1{,}208 & 1{,}056 & 2{,}139 & 1{,}296 & 1{,}112 \\
$N_{\mathrm{failed\_first}}$ & Failed first attempts & 435 & 365 & 351 & 527 & 391 & 355 \\
$N_{\mathrm{retry\_succ}}$ & Successful retry corrections & 166 & 30 & 9 & 268 & 55 & 17 \\
\bottomrule
\end{tabular}
\end{table}

\subsection{Compute resources}
\label[appendix]{appendix:compute-resources}

The three LLM controllers were accessed directly through their respective APIs, and no
model training, fine-tuning, or local GPU inference was performed.
The main computational cost came from LLM command-generation interactions and
command processing in the simulator.
Summed over the three controllers, the evaluation included 4{,}211 attempted target transitions in total: 1{,}980 for Task 1 and
2{,}231 for Task 2, as reported in~\cref{tab:appendix-metric-counts}.
Including retry attempts, the experiments involved 8{,}699 command attempts in total:
4{,}152 for Task 1 and 4{,}547 for Task 2.
Because each command attempt required an LLM command-generation output, 8{,}699 is a
lower bound on the number of LLM command-generation interactions; this excludes initial
context-upload turns and auxiliary setup interactions.
The total time recorded by evaluators in the evaluation interface, summed over the three controllers, was approximately 2{,}407 minutes (about 40 hours) for
Task 1 and 8{,}658 minutes (about 144 hours) for Task 2.
This includes waiting for model responses and, in Task 2, the time the instructor spent writing guidance and answering clarification questions.

\subsection{Human-guided correction analysis}
\label[appendix]{appendix:human-guided-correction-analysis}

The main paper reports that Retry-SR is higher in Task 2 than in Task 1 for all three controllers
(GPT 5.6 Sol: 38.2\% to 50.9\%; Gemma 4 31B: 8.2\% to 14.1\%; Gemma 3 12B IT: 2.6\% to 4.8\%).
This section provides additional interpretation of this descriptive result.

Origami state transitions often require information about orientation, visible and hidden
layers, reference points, and the intended local operation.
In Task 1, fixed feedback only indicates that the command is invalid or that the fold is
incorrect, which may be insufficient for identifying which part of the command should be
revised.
In Task 2, a human instructor can provide targeted corrective information, such as which
layer should be folded, which visible reference point should be used, or how the intended
operation differs from the failed command.

This result supports the usefulness of Task 2 as an instructor-mediated state-transition
benchmark.
However, it should be interpreted cautiously.
The simulator keeps the current state image unchanged after a failed attempt, so retry-based
correction does not measure full physical recovery from a deviated folded state.
Task 2 performance may also depend on instructor phrasing, corrective-feedback style,
clarification behavior, and uneven instructor workload.
Thus, the result provides preliminary evidence that human guidance can improve retry-based
command correction, but not a fully controlled evaluation of collaborative physical recovery.

\end{document}